\PassOptionsToPackage{unicode}{hyperref}
\PassOptionsToPackage{hyphens}{url}
\documentclass[
  11pt,
]{article}
\usepackage{amsmath,amssymb}
\usepackage{iftex}
\ifPDFTeX
  \usepackage[T1]{fontenc}
  \usepackage[utf8]{inputenc}
  \usepackage{textcomp} 
\else 
  \usepackage{unicode-math} 
  \defaultfontfeatures{Scale=MatchLowercase}
  \defaultfontfeatures[\rmfamily]{Ligatures=TeX,Scale=1}
\fi
\ifPDFTeX\else
\fi
\IfFileExists{upquote.sty}{\usepackage{upquote}}{}
\IfFileExists{microtype.sty}{
}{}
\makeatletter
\@ifundefined{KOMAClassName}{
  \IfFileExists{parskip.sty}{%
    \usepackage{parskip}
  }{
    \setlength{\parindent}{0pt}
    \setlength{\parskip}{6pt plus 2pt minus 1pt}}
}{
  \KOMAoptions{parskip=half}}
\makeatother
\usepackage{xcolor}
\usepackage[margin=1in]{geometry}
\usepackage{longtable,booktabs,array}
\usepackage{calc} 
\usepackage{etoolbox}
\makeatletter
\patchcmd\longtable{\par}{\if@noskipsec\mbox{}\fi\par}{}{}
\makeatother
\IfFileExists{footnotehyper.sty}{\usepackage{footnotehyper}}{\usepackage{footnote}}
\makesavenoteenv{longtable}
\usepackage{amsthm}

\theoremstyle{remark}

\usepackage{enumitem}

\ifLuaTeX
\fi
\usepackage[]{natbib}
\IfFileExists{bookmark.sty}{\usepackage{bookmark}}{\usepackage{hyperref}}
\IfFileExists{xurl.sty}{\usepackage{xurl}}{} 
\hypersetup{
  pdftitle={The Digital Twin Counterfactual Framework: A Validation Architecture for Simulated Potential Outcomes},
  pdfauthor={Olav Laudy},
  hidelinks,
  pdfcreator={LaTeX via pandoc}}

\title{The Digital Twin Counterfactual Framework: A Validation
Architecture for Simulated Potential Outcomes}
\author{Olav Laudy}
\date{}

\begin{document}
\maketitle
\begin{abstract}
The fundamental problem of causal inference --- that the counterfactual
outcome for any individual is never observed --- has shaped the entire
methodology of the field. Every existing approach addresses this problem
by substituting assumptions for missing data: ignorability, parallel
trends, exclusion restrictions. None produces the counterfactual itself.
This paper proposes the \textbf{Digital Twin Counterfactual Framework
(DTCF)}, which takes a different approach: rather than estimating the
counterfactual statistically, we \emph{simulate} it using a digital twin
--- a computational replica of each individual placed under both
treatment and control conditions --- and then subject the simulation to
a hierarchical validation regime that makes explicit which causal
quantities are empirically disciplined, which remain
assumption-dependent, and how large the residual uncertainty is. We
formalize the digital twin simulator as a stochastic mapping within the
potential outcomes framework and introduce a hierarchy of \emph{twin
fidelity assumptions} --- from marginal fidelity through joint fidelity
to structural fidelity --- each unlocking a progressively richer class
of estimands. The central methodological contribution is threefold.
First, a five-level \emph{validation architecture} converts the
unfalsifiable claim ``the simulator produces correct counterfactuals''
into falsifiable tests against observable data, establishing which
estimands are credible at each validation level. Second, a formal
decomposition separates causal quantities into those that are
\emph{marginally validated} (ATE, CATE, QTE --- testable through
observable-arm comparison) and those that are \emph{copula-dependent}
(the ITE distribution, probability of benefit/harm, variance of
treatment effects --- permanently reliant on the unobservable
within-individual dependence structure). Third, a suite of bounding,
sensitivity, and uncertainty quantification tools makes the copula
dependence explicit rather than hidden. The DTCF does not resolve the
fundamental problem of causal inference --- the counterfactual remains
unobserved. What it provides is a framework in which marginal causal
claims become increasingly testable, joint causal claims become
explicitly assumption-indexed, and the gap between the two is formally
characterized. The framework applies to any simulation technology; we
discuss implementation with large language models as one candidate
engine, noting the substantial open challenges.
\end{abstract}

\hypertarget{introduction}{%
\section{Introduction}\label{introduction}}

\hypertarget{the-fundamental-problem}{%
\subsection{The Fundamental
Problem}\label{the-fundamental-problem}}

Causal inference rests on a deceptively simple comparison: what happened
versus what \emph{would have} happened. In the potential outcomes
framework (\citet{neyman1923application}; \citet{rubin1974estimating}), every individual \(i\) possesses
two potential outcomes --- \(Y_i(1)\) under treatment and \(Y_i(0)\)
under control --- and the individual treatment effect is
\(\tau_i = Y_i(1) - Y_i(0)\). The fundamental problem (\citet{holland1986statistics}) is
that we can never observe both. Each individual is either treated or
untreated, never both. The counterfactual is permanently missing.

This missing data problem has shaped the architecture of modern causal
inference. Randomized controlled trials solve it at the population level
by ensuring exchangeability between groups. When randomization is
infeasible, observational methods --- matching, inverse probability
weighting, instrumental variables, difference-in-differences, regression
discontinuity, synthetic controls --- substitute assumptions for the
missing data \citep{rosenbaum1983central, angrist1996identification, abadie2010synthetic}. Each requires ignorability,
exclusion restrictions, parallel trends, or continuity at a threshold.
None produces the counterfactual itself.

The consequence is a fundamental asymmetry in what causal inference can
deliver. The ATE is popular not because it is the most useful quantity,
but because it is the most \emph{estimable}. The individual treatment
effect --- what clinicians and policymakers actually need --- remains
out of reach. The full distribution of treatment effects, its variance,
and the probability that treatment harms rather than helps, are
inaccessible without observing both potential outcomes.

\hypertarget{the-digital-twin-as-a-candidate-substitute}{%
\subsection{The Digital Twin as a Candidate
Substitute}\label{the-digital-twin-as-a-candidate-substitute}}

Digital twins --- computational replicas of real-world entities --- have
migrated from engineering \citep{grieves2014digital} into healthcare
\citep{corralacero2020digital, bjornsson2020digital} and social science.
A parallel development has demonstrated that large language models can
simulate human behavioral responses with non-trivial fidelity: \citet{argyle2023silicon} replicated survey patterns via ``silicon sampling,'' a
mega-study \citep{peng2025funhouse} found individual-level prediction accuracy of
approximately 0.75 across diverse behavioral domains, and \citet{koaik2026social} demonstrated calibration of LLM-based ``Social Digital
Twins'' against real-world mobility data. In healthcare, \citet{vallee2026prediction}
proposed integrating structural causal models with digital twins, while
\citet{qian2021synctwin} developed SyncTwin for constructing synthetic
counterfactuals from observational data.

\hypertarget{what-is-missing}{%
\subsection{What Is Missing}\label{what-is-missing}}

Despite this convergence, the existing literature treats digital twins
as predictive tools (forecasting outcomes, not generating
counterfactuals), applies causal methods \emph{to} twins (using
causality to validate twins, not the reverse), or constructs synthetic
twins from weighted composites of other individuals (estimating the
counterfactual statistically rather than simulating it independently).

A skeptical reader may observe that what we propose --- simulate
outcomes under interventions and trust the simulation if it predicts
well --- is just a restatement of the model-based or structural
generative modeling agenda. We acknowledge the family resemblance but
contend that the contribution is more specific. What has not been
provided in the existing literature is: (1) a formal embedding of the
generative simulator within the potential outcomes framework as a direct
stand-in for the missing counterfactual; (2) a decomposition of causal
credibility into \emph{progressively testable fidelity layers}, with a
precise mapping from each layer to the estimands it licenses; and (3) a
formal accounting of what remains \emph{permanently untestable} ---
specifically, the copula between potential outcomes --- with bounding
and sensitivity tools that make this gap explicit and quantifiable
rather than implicit and ignored.

The core novelty, stated plainly, is not the use of a simulator for
causal inference. It is the architecture that separates what
simulation-plus-validation can discipline from what it cannot, and the
tools for managing the residual gap.

At a conceptual level, the DTCF replaces the classical strategy of
inferring missing potential outcomes from other individuals with a
fundamentally different approach: for each individual, we construct a
digital twin that generates outcomes under both treatment and control.
The resulting simulated pair $(\hat{Y}_i(1), \hat{Y}_i(0))$ serves as a
proxy for the unobservable potential outcomes $(Y_i(1), Y_i(0))$. This
does not reveal the true counterfactual; instead, it produces a
model-based substitute whose credibility depends on how well the
simulator reproduces observable outcomes. The role of the validation
architecture developed in this paper is to determine which causal
quantities derived from these simulated paired potential outcomes are
empirically supported and which remain assumption-dependent.

\hypertarget{contribution}{%
\subsection{Contribution}\label{contribution}}

This paper proposes the \textbf{Digital Twin Counterfactual Framework
(DTCF)}. We formalize the digital twin simulator as a stochastic mapping
(Section 2), introduce a hierarchy of fidelity assumptions mapped to the
estimands each identifies, develop a five-level validation architecture
with an explicit theory of when factual validation transfers to
counterfactual credibility --- and when it does not (Section 3),
enumerate the taxonomy of computable estimands with an honest accounting
of what each requires (Section 4), and confront the irreducible
limitation --- the joint distribution problem --- with bounding,
sensitivity, and uncertainty quantification (Section 5). Section 6
briefly discusses LLM-based implementation as one candidate engine, and
Section 7 concludes.

The central contribution, in one sentence: \textbf{the DTCF decomposes
causal inference into a marginal component that can be empirically
validated against observable data, and a joint component that is
provably untestable but can be bounded, sensitivity-analyzed, and
explicitly reported.} This is a different kind of advance than classical
identification results, and a more defensible one than claiming the
fundamental problem is solved.

\hypertarget{the-dtcf-framework-and-assumptions}{%
\section{The DTCF: Framework and
Assumptions}\label{the-dtcf-framework-and-assumptions}}

\hypertarget{setup}{%
\subsection{Setup}\label{setup}}

We adopt the standard potential outcomes framework. Each individual
\(i\) in a population \(\mathcal{P} = \{1, \ldots, N\}\) has covariates
\(X_i \in \mathcal{X}\), treatment indicator \(D_i \in \{0,1\}\), and
potential outcomes \((Y_i(1), Y_i(0))\). Under SUTVA (\citet{rubin1980randomization}), the
observed outcome is \(Y_i^{\text{obs}} = D_i Y_i(1) + (1-D_i)Y_i(0)\),
and the individual treatment effect \(\tau_i = Y_i(1) - Y_i(0)\) is
never observable. Classical identification of even population-level
estimands --- the ATE, ATT, CATE --- requires strong ignorability:
\(\{Y_i(1), Y_i(0)\} \perp\!\!\!\perp D_i \mid X_i\) with overlap. Even
then, the joint distribution of \((Y_i(1), Y_i(0))\) --- and therefore
the ITE distribution, \(\Pr(\tau_i > 0)\), and \(\text{Var}(\tau_i)\)
--- remains unidentified.

\hypertarget{the-digital-twin-simulator}{%
\subsection{The Digital Twin
Simulator}\label{the-digital-twin-simulator}}

\textbf{Definition 1 (Digital Twin Simulator).} A digital twin simulator
is a stochastic mapping
\(\mathcal{S}: \mathcal{X} \times \{0,1\} \times \mathcal{U} \to \mathcal{Y}\),
where \(\mathcal{U}\) is a latent noise space with measure \(\mu\). For
individual \(i\) with covariates \(X_i\), the simulated potential
outcome under treatment \(d\) is:

\[\hat{Y}_i(d) = \mathcal{S}(X_i, d, U_i), \quad U_i \sim \mu.\]

The simulator induces conditional distributions
\(\hat{F}_d(y \mid x) = \Pr(\hat{Y}_i(d) \leq y \mid X_i = x)\), to be
compared against the real-world conditionals
\(F_d(y \mid x) = \Pr(Y_i(d) \leq y \mid X_i = x)\).

\textbf{Definition 2 (Digital Twin Counterfactual).} For individual
\(i\) assigned to \(D_i = 1\), the digital twin counterfactual is
\(\hat{Y}_i^{\text{cf}} = \mathcal{S}(X_i, 0, U_i)\); symmetrically for
\(D_i = 0\). The simulated ITE is:

\[\hat{\tau}_i = \mathcal{S}(X_i, 1, U_i) - \mathcal{S}(X_i, 0, U_i),\]

where crucially the \emph{same} noise draw \(U_i\) is used for both
potential outcomes. This shared coupling generates the joint
distribution \((\hat{Y}_i(1), \hat{Y}_i(0))\) and is the mechanism
through which individual-level quantities become computable. A
clarification is essential here: the simulator does not reveal the true
counterfactual. It provides a model-based proxy whose credibility is
determined entirely by validation against observable quantities and by
the plausibility of its assumptions. If independent noise draws were
used, \(\text{Corr}(\hat{Y}_i(1), \hat{Y}_i(0) \mid X_i)\) would be
forced to zero, artificially inflating ITE variance. The shared \(U_i\)
encodes the simulator's model of within-individual dependence --- a
quantity fundamentally unobservable in the real world. We return to the
implications of this encoding --- and the difficulty of knowing whether
it is correct --- in Sections 3 and 5.

\hypertarget{twin-fidelity-assumptions}{%
\subsection{Twin Fidelity
Assumptions}\label{twin-fidelity-assumptions}}

Identification in the DTCF replaces the classical ignorability
assumption with a fidelity assumption about the simulator. This is not a
free lunch --- it trades one class of untestable assumptions for
another, with the advantage that the new assumptions are
\emph{partially} testable.

\textbf{Assumption 1 (Twin Fidelity --- Strong Form).} The simulator
satisfies \emph{strong twin fidelity} if
\(\hat{F}_d(\cdot \mid x) = F_d(\cdot \mid x)\) for all
\(d \in \{0,1\}\) and all \(x \in \mathcal{X}\).

Under strong twin fidelity, all estimands depending on the marginal
distributions separately --- ATE, ATT, ATU, CATE, QTE --- are exactly
identified. However, strong twin fidelity does \emph{not} guarantee
correctness of the ITE distribution, because it constrains only the
marginals, not their joint. This is the central tension in the
framework.

\textbf{Proposition 1 (Sklar Decomposition).} By Sklar's theorem, the
joint conditional distribution decomposes as
\(H(y_1, y_0 \mid x) = C_x(F_1(y_1 \mid x), F_0(y_0 \mid x))\), where
\(C_x\) is a copula capturing dependence between potential outcomes. The
simulator induces
\(\hat{H}(y_1, y_0 \mid x) = \hat{C}_x(\hat{F}_1(y_1 \mid x), \hat{F}_0(y_0 \mid x))\).

\textbf{Corollary 1.} Strong twin fidelity guarantees
\(\hat{F}_d = F_d\) but does \emph{not} constrain \(\hat{C}_x\) relative
to \(C_x\). The simulator's joint distribution matches reality if and
only if both strong twin fidelity holds \emph{and} \(\hat{C}_x = C_x\)
for all \(x\).

This is the irreducible gap. The marginals are testable; the copula is
not. Every individual-level causal quantity in the DTCF lives on the
copula side of this divide.

\textbf{Assumption 2 (Twin Fidelity --- Joint Form).} The simulator
satisfies \emph{joint twin fidelity} if
\(\hat{H}(\cdot, \cdot \mid x) = H(\cdot, \cdot \mid x)\) for all \(x\)
--- equivalently, both strong twin fidelity and \(\hat{C}_x = C_x\).
Joint twin fidelity is strictly stronger and requires the simulator to
model how an individual's outcomes \emph{co-vary} across conditions.
This component can never be directly validated from observational data.
The best we can do is bound, constrain, and stress-test it (Section 5).

\hypertarget{approximate-fidelity-and-error-propagation}{%
\subsection{Approximate Fidelity and Error
Propagation}\label{approximate-fidelity-and-error-propagation}}

No simulator achieves exact fidelity. We develop the framework under
approximation.

\textbf{Definition 3 (\(\varepsilon\)-Fidelity).} The simulator
satisfies \(\varepsilon\)-fidelity if
\(\sup_{x,d} d_{\text{KS}}(\hat{F}_d(\cdot \mid x), F_d(\cdot \mid x)) \leq \varepsilon\),
where \(d_{\text{KS}}\) is the Kolmogorov-Smirnov distance. Note
\(\varepsilon = 0\) recovers Assumption 1.

\textbf{Theorem 1 (ATE Error Bound).} Under \(\varepsilon\)-fidelity
with bounded outcomes \(Y_i(d) \in [a,b]\):
\(|\widehat{\text{ATE}} - \text{ATE}| \leq 2\varepsilon(b-a)\).

The bound extends identically to the CATE pointwise at each \(x\)
(Theorem 2 in Appendix B). For the ITE distribution, a stronger
condition is needed:

\textbf{Theorem 3 (ITE Distribution Error).} Under joint
\(\varepsilon_J\)-fidelity ---
\(\sup_x \sup_{(y_1,y_0)} |\hat{H}(y_1,y_0 \mid x) - H(y_1,y_0 \mid x)| \leq \varepsilon_J\)
--- the ITE distribution satisfies
\(\sup_t |\hat{G}_\tau(t \mid x) - G_\tau(t \mid x)| \leq \varepsilon_J\).

The critical distinction: ATE and CATE depend only on marginal fidelity;
the ITE distribution requires joint fidelity incorporating the copula.
This asymmetry structures everything that follows.

\hypertarget{the-22-structure}{%
\subsection{The 2×2 Structure}\label{the-22-structure}}

The DTCF operates on a structural insight. For each individual, the
simulation produces outcomes under both conditions. The real world
produces outcomes under only one:

\begin{longtable}[]{@{}
  >{\raggedright\arraybackslash}p{(\columnwidth - 4\tabcolsep) * \real{0.1782}}
  >{\raggedright\arraybackslash}p{(\columnwidth - 4\tabcolsep) * \real{0.4158}}
  >{\raggedright\arraybackslash}p{(\columnwidth - 4\tabcolsep) * \real{0.4059}}@{}}
\toprule\noalign{}
\begin{minipage}[b]{\linewidth}\raggedright
\end{minipage} & \begin{minipage}[b]{\linewidth}\raggedright
Treatment arm (\(d=1\))
\end{minipage} & \begin{minipage}[b]{\linewidth}\raggedright
Control arm (\(d=0\))
\end{minipage} \\
\midrule\noalign{}
\endhead
\bottomrule\noalign{}
\endlastfoot
\textbf{Simulated} & \(\{\hat{Y}_i(1)\}_{i=1}^N\) &
\(\{\hat{Y}_i(0)\}_{i=1}^N\) \\
\textbf{Real-world} & \(\{Y_i(1)\}_{i \in \mathcal{P}_1}\) &
\(\{Y_i(0)\}_{i \in \mathcal{P}_0}\) \\
\end{longtable}

The \emph{diagonal} comparisons --- simulated treatment vs.~real
treatment, simulated control vs.~real control --- are the basis of
validation. The \emph{off-diagonal} entries are the counterfactuals of
inferential interest. Under \(\varepsilon\)-fidelity, the ATE, ATT, ATU,
CATE, and QTE are all identified with bias bounded by
\(2\varepsilon(b-a)\) (Proposition 2; proof in Appendix B). The ITE
distribution, probability of benefit/harm, ITE variance, and probability
of causation require joint fidelity (Proposition 3; Appendix B).

\hypertarget{extensions-mediation-and-dynamic-regimes}{%
\subsection{Extensions: Mediation and Dynamic
Regimes}\label{extensions-mediation-and-dynamic-regimes}}

The DTCF extends, in principle, to two settings that are classically
challenging. We describe the extensions here with the caveat that both
require strong fidelity conditions that are difficult to verify, and the
claims should be read as ``possible under stated assumptions'' rather
than ``achieved.''

\textbf{Mediation.} Natural direct and indirect effects require
\emph{cross-world counterfactuals} --- e.g., \(Y_i(1, M_i(0))\).
Classical identification demands cross-world independence \citep{pearl2001direct}
or sequential ignorability \citep{imai2010identification}. In the DTCF,
the simulator directly produces cross-world quantities by composition,
replacing these classical assumptions with \textbf{Assumption 3
(Structural Fidelity)}: the simulator correctly models the conditional
distributions along causal paths --- \(\hat{F}_{Y|D,M,X} = F_{Y|D,M,X}\)
and \(\hat{F}_{M|D,X} = F_{M|D,X}\). This is a \emph{different} strong
assumption, not a weaker one. Structural fidelity asks whether the
simulator has correctly internalized the causal mechanism
\(D \to M \to Y\). For simulators built from observational patterns
(such as LLMs), this is an extremely demanding requirement, because the
simulator must have learned not just associations but the interventional
conditional distributions. Whether any existing simulator satisfies
structural fidelity for a given mediation problem is an empirical
question that the validation architecture can partially, but not fully,
address --- Level 1 can be applied to each structural component, but the
cross-world composition introduces untestable joint assumptions
analogous to the copula problem.

\textbf{Dynamic treatment regimes.} For a \(T\)-period setting, the
simulator can be rolled forward under any candidate regime, yielding
direct evaluation and optimization. This requires \textbf{Assumption 4
(Sequential Twin Fidelity)}: correct one-step-ahead conditional
distributions at each time point. This is demanding for long horizons,
as errors may compound even if each step is individually
well-calibrated.

\hypertarget{summary-assumptions-and-their-implications}{%
\subsection{Summary: Assumptions and Their
Implications}\label{summary-assumptions-and-their-implications}}

\begin{longtable}[]{@{}
  >{\raggedright\arraybackslash}p{(\columnwidth - 6\tabcolsep) * \real{0.2500}}
  >{\raggedright\arraybackslash}p{(\columnwidth - 6\tabcolsep) * \real{0.2500}}
  >{\raggedright\arraybackslash}p{(\columnwidth - 6\tabcolsep) * \real{0.2500}}
  >{\raggedright\arraybackslash}p{(\columnwidth - 6\tabcolsep) * \real{0.2500}}@{}}
\toprule\noalign{}
\begin{minipage}[b]{\linewidth}\raggedright
Assumption
\end{minipage} & \begin{minipage}[b]{\linewidth}\raggedright
Formal Condition
\end{minipage} & \begin{minipage}[b]{\linewidth}\raggedright
Estimands Identified
\end{minipage} & \begin{minipage}[b]{\linewidth}\raggedright
Testability
\end{minipage} \\
\midrule\noalign{}
\endhead
\bottomrule\noalign{}
\endlastfoot
SUTVA & No interference, no hidden treatment variants & All (baseline) &
Untestable \\
\(\varepsilon\)-Fidelity (Def. 3) &
\(d_{\text{KS}}(\hat{F}_d(\cdot \mid x), F_d(\cdot \mid x)) \leq \varepsilon\)
& ATE, ATT, ATU, CATE, QTE & \textbf{Testable} (Levels 0--3) \\
Joint Twin Fidelity (Asm. 2) &
\(\hat{H}(\cdot,\cdot \mid x) = H(\cdot,\cdot \mid x)\) & ITE
distribution, \(\pi_+\), \(\pi_-\), Var(\(\tau_i\)), PC &
\textbf{Marginals testable; copula not} \\
Structural Fidelity (Asm. 3) & Correct conditionals along causal paths &
NDE, NIE, path-specific effects & Partially testable per component \\
Sequential Twin Fidelity (Asm. 4) & Correct one-step-ahead conditionals
over time & DTR values, optimal regime identification & Testable per
step; composition untestable \\
\end{longtable}

The rightmost column --- absent from earlier drafts --- is the point.
The DTCF's contribution is not that all assumptions become testable. It
is that the \emph{testable portion is explicitly separated from the
untestable portion}, and the untestable portion is bounded and
sensitivity-analyzed rather than buried.

\hypertarget{the-validation-architecture}{%
\section{The Validation
Architecture}\label{the-validation-architecture}}

\hypertarget{the-validation-principle-and-its-limits}{%
\subsection{The Validation Principle and Its
Limits}\label{the-validation-principle-and-its-limits}}

The central methodological contribution of the DTCF is that --- unlike
classical assumptions (ignorability, parallel trends, exclusion
restrictions) --- the core fidelity assumptions are \emph{partially
testable}.

\textbf{Principle 1 (Observable-Arm Validation).} The counterfactual arm
cannot be directly validated. However, the \emph{factual} arm --- the
simulated outcome under the condition actually administered --- can be
compared against the observed outcome. If the simulator is accurate
where it can be checked, its accuracy where it cannot be checked is
rendered more credible.

The logical structure is: (1) the simulator uses the same mechanism
\(\mathcal{S}\) for both conditions; (2) we test whether \(\mathcal{S}\)
is accurate under the observed condition; (3) if it is, and there is no
reason to believe \(\mathcal{S}\) behaves differently under the
counterfactual condition, the counterfactual outputs inherit
credibility.

Step 3 is not automatic. It requires an explicit assumption that carries
substantial weight in the framework.

\hypertarget{transportability-across-arms}{%
\subsection{Transportability Across
Arms}\label{transportability-across-arms}}

\textbf{Assumption 5 (Transportability Across Arms).} If the simulator
satisfies \(\varepsilon\)-fidelity for arm \(d\) at covariate \(x\),
then fidelity for \(d' \neq d\) at the same \(x\) is bounded by:

\[d_{\text{KS}}(\hat{F}_{d'}(\cdot \mid x), F_{d'}(\cdot \mid x)) \leq \varepsilon + \delta(x, d, d'),\]

where \(\delta(x, d, d') \geq 0\) is the extrapolation penalty.

This assumption is doing major work and deserves careful treatment. It
is the bridge from factual validation to counterfactual credibility ---
the analogue of what overlap and ignorability do in classical causal
inference. We are transparent that it carries similar risks: it is
indispensable, only partially verifiable, and can fail silently.

\textbf{When transportability is plausible.} The penalty \(\delta\) is
expected to be small when: (a) the treatment and control conditions are
\emph{structurally similar} --- they involve variations within the same
domain rather than categorically different interventions (e.g., two drug
doses vs.~drug vs.~surgery); (b) the simulator's mechanism is
\emph{domain-general} --- it operates on the same internal model for
both arms rather than having specialized, separately-trained components
for treatment and control; (c) both arms have \emph{some} representation
in the validation data --- even if the split is unequal, both arms are
partially validated, bounding \(\delta\) from both sides.

\textbf{When transportability is suspect.} The penalty \(\delta\) may be
large when: (a) the treatment is \emph{novel} relative to the
simulator's training domain --- if the simulator has never encountered
anything like the treatment condition, its extrapolation from the
control arm is speculative; (b) treatment and control activate
\emph{different mechanisms} --- e.g., a behavioral intervention vs.~a
pharmacological one, where the simulator may have learned one well and
the other poorly; (c) one arm is validated and the other is not --- the
fully off-diagonal case, where all counterfactual credibility rests on
transport.

\textbf{Diagnosing transportability.} We propose three practical checks:
1. \emph{Arm-specific validation discrepancies.} If the simulator is
validated separately on subsets assigned to each arm (possible in
observational data where both treated and control individuals exist),
compare the arm-specific \(\varepsilon\) values. A large discrepancy
\(|\varepsilon_1 - \varepsilon_0|\) is direct evidence of differential
fidelity, and the larger value provides a lower bound on \(\delta\). 2.
\emph{Treatment effect calibration (Level 3).} If an RCT or
quasi-experimental benchmark is available, the ATE discrepancy
\(T^{(3)}\) provides an aggregate test of whether transport-induced bias
is consequential. This does not isolate \(\delta\), but a large
\(T^{(3)}\) with small arm-specific \(\varepsilon\) values is a
signature of transport failure. 3. \emph{Placebo transport tests.}
Evaluate the simulator under two conditions that are known to produce
the same outcome distribution (e.g., two identical placebos, or two time
periods with the same intervention). If the simulator shows differential
performance across these ``equivalent arms,'' transportability is
compromised.

\textbf{Reporting recommendation.} Any DTCF analysis should report: (i)
the arm-specific \(\varepsilon\) values separately; (ii) the magnitude
of the ATE discrepancy from Level 3 when available; and (iii) a
qualitative assessment of whether the treatment and control conditions
are within the simulator's extrapolation envelope. When \(\delta\) is
judged to be non-negligible, the error bounds on all estimands should be
widened accordingly:
\(|\widehat{\text{ATE}} - \text{ATE}| \leq (\varepsilon_1 + \delta_1)(b-a) + (\varepsilon_0 + \delta_0)(b-a)\).

\hypertarget{the-validation-hierarchy}{%
\subsection{The Validation
Hierarchy}\label{the-validation-hierarchy}}

We define five levels of validation, each testing a progressively
stronger fidelity condition and licensing a richer set of estimands. The
levels are cumulative.

\textbf{Level 0: Marginal Calibration.} Tests whether the unconditional
distribution of simulated outcomes matches reality, separately per arm
--- using, e.g., two-sample Kolmogorov-Smirnov tests. Passing Level 0
establishes marginal distributional match but permits errors that cancel
across subgroups. It licenses the marginal ATE with the caveat that
subgroup effects may be wrong.

\textbf{Level 1: Conditional Calibration.} Tests whether the simulated
outcome distribution matches reality \emph{within covariate strata} ---
either by stratification with multiplicity-corrected KS tests or by
conditional maximum mean discrepancy for continuous covariates \citep{park2020measure}. This is the critical threshold for population-level and
subgroup-level estimands.

\textbf{Theorem 4 (Estimand Credibility at Level 1).} If Level 1
validation passes at resolution \(\varepsilon_1\), then for bounded
outcomes in \([a,b]\): (a)
\(|\widehat{\text{ATE}} - \text{ATE}| \leq 2\varepsilon_1(b-a)\); (b)
same bound for ATT, ATU; (c)
\(|\widehat{\text{CATE}}(x) - \text{CATE}(x)| \leq 2\varepsilon_1(b-a)\)
within validated strata; (d) same bound for QTE. The key point:
\(\varepsilon_1\) is \emph{measured}, not postulated. These bounds hold
\emph{on the observable arms}; the additional uncertainty from transport
to the counterfactual arm enters through \(\delta\) per Assumption 5.

\textbf{Level 2: Individual-Level Calibration.} Tests whether the
simulator accurately predicts each individual's \emph{observed} outcome,
via RMSPE, calibration slope/intercept (\(\beta_0 = 0, \beta_1 = 1\)
implies perfect calibration), and conditional coverage of prediction
intervals. A simulator can pass Level 1 while failing Level 2: it may
get conditional distributions right but assign the wrong outcome to the
wrong individual within a stratum.

Level 2 provides the strongest observable evidence for the
counterfactual. If the factual prediction error
\(e_i = Y_i^{\text{obs}} - \hat{Y}_i(D_i)\) is small and the
extrapolation penalty \(\delta\) is limited, then counterfactual
predictions are also accurate --- formally,
\(\mathbb{E}[(Y_i(1-D_i) - \hat{Y}_i(1-D_i))^2] \leq \mathbb{E}[e_i^2] + \Delta_{\text{transport}}\)
(Theorem 5; proof in Appendix B). This implication is conditional on
transportability (Assumption 5); without it, individual-level factual
accuracy does not guarantee counterfactual accuracy --- a simulator can
predict every observed outcome perfectly while generating arbitrarily
wrong counterfactuals if the mechanism behaves differently across arms.
The quantity \(\Delta_{\text{transport}}\) is not directly measurable;
it depends on \(\delta\), which must be assessed by the methods of
Section 3.2.

\textbf{Level 3: Treatment Effect Calibration.} Tests whether the
simulator's estimated treatment effects match effects from an RCT or
credible quasi-experiment. Given a randomized subset \(\mathcal{R}\),
the discrepancy
\(T^{(3)} = \widehat{\text{ATE}}_{\text{sim}} - \widehat{\text{ATE}}_{\text{RCT}}\)
is tested via bootstrap. This is the headline validation --- it directly
compares the simulator's primary causal output against the gold
standard. Extension to CATE calibration across strata yields a
calibration plot, where perfect calibration lies on the 45-degree line.

Level 3 is also the most informative diagnostic for transportability: if
the simulator passes Levels 0--2 on observable arms but fails Level 3 on
treatment effects, the discrepancy is evidence that factual accuracy is
not transferring to counterfactual accuracy --- i.e., that \(\delta\) is
non-negligible.

\textbf{Level 4: Distributional Stress Testing.} A suite of tests
probing the copula and simulator robustness: (a) \emph{copula
sensitivity} --- re-run with independent noise to produce the
independence-copula ITE distribution and compute the copula sensitivity
index
\(\text{CSI} = d_{\text{KS}}(\hat{G}_\tau, \hat{G}_\tau^{\text{ind}})\);
(b) \emph{Fréchet-Hoeffding bounds} on ITE-dependent quantities; (c)
dose-response monotonicity checks; (d) placebo tests under conditions
where the true effect is known to be zero.

Full statistical details for all levels --- test statistics, asymptotic
distributions, multiplicity corrections, sample size requirements, and
the integrated validation protocol --- are provided in Appendix C.

\hypertarget{the-limits-of-validation}{%
\subsection{The Limits of
Validation}\label{the-limits-of-validation}}

\textbf{Theorem 7 (Validation Completeness).} (a) All estimands
depending only on the marginal distributions \(F_1(\cdot \mid x)\) and
\(F_0(\cdot \mid x)\) --- ATE, ATT, ATU, CATE, QTE --- can be validated
to arbitrary precision through Levels 0--3. (b) All estimands depending
on the copula \(C_x\) --- ITE distribution, \(\Pr(\tau_i > 0)\),
\(\text{Var}(\tau_i)\), probability of causation --- cannot be validated
from observable data. They can only be bounded, stress-tested, or
assumed correct.

This dichotomy is the framework's core structural result, and it maps
directly to the testability column in Section 2.7. The DTCF's response
to part (b) --- bounding, sensitivity analysis, and explicit uncertainty
quantification for copula-dependent quantities --- is developed in
Section 5.

\hypertarget{what-becomes-computable}{%
\section{What Becomes Computable}\label{what-becomes-computable}}

The DTCF renders three categories of estimands computable. The first
comprises classically identifiable quantities --- ATE, ATT, ATU, CATE,
QTE --- for which the DTCF replaces untestable assumptions
(ignorability, overlap) with the partially testable twin fidelity
assumption (plus transportability). The second, and the primary
motivation for the framework, comprises previously inaccessible
individual-level and distributional quantities --- which become
available only under the additional (untestable) assumption of correct
copula specification. The third comprises structurally complex
quantities --- mediation, dynamic regimes, spillovers --- that
classically require very strong assumptions and in the DTCF require
comparably strong fidelity conditions.

We illustrate with three key examples; the complete catalog for all 17
estimands is in Appendix D.

\hypertarget{the-ate-under-the-dtcf}{%
\subsection{The ATE Under the
DTCF}\label{the-ate-under-the-dtcf}}

The simplest case. The DTCF estimator is
\(\widehat{\text{ATE}} = N^{-1}\sum_i (\hat{Y}_i(1) - \hat{Y}_i(0))\),
requiring only \(\varepsilon\)-fidelity (validated at Level 1, confirmed
at Level 3). The error bound is \(2\varepsilon_1(b-a)\), widened by the
transport penalty \(\delta\) for the counterfactual arm. No ignorability
or overlap is needed --- but twin fidelity and transportability are
needed instead. This is an assumption trade, not an assumption
elimination.

\hypertarget{the-ite-distribution-and-probability-of-benefit}{%
\subsection{The ITE Distribution and Probability of
Benefit}\label{the-ite-distribution-and-probability-of-benefit}}

These are the estimands that most strongly motivate the DTCF --- and the
ones most dependent on the untestable copula.

\textbf{Individual Treatment Effect.} The DTCF estimator is
\(\hat{\tau}_i = \hat{Y}_i(1) - \hat{Y}_i(0)\), requiring joint twin
fidelity. Classical causal inference cannot identify the ITE; the DTCF
produces a point estimate whose accuracy depends on the copula (the
untestable component per Theorem 7).

\textbf{ITE Distribution.} The CDF
\(\hat{G}_\tau(t) = N^{-1}\sum_i \mathbf{1}(\hat{\tau}_i \leq t)\)
yields the full shape of treatment effect heterogeneity. From it we
compute: the variance \(\widehat{\text{Var}}(\tau_i)\); skewness and
higher moments; and multimodality --- which indicates distinct
subpopulations with qualitatively different treatment responses.
\citet{heckman1997making} showed this distribution is
unidentified from experimental data alone; the DTCF produces a point
estimate under joint fidelity together with Fréchet-Hoeffding bounds as
a robustness check.

\textbf{Probability of Benefit and Harm.} The estimators
\(\hat{\pi}_+ = N^{-1}\sum_i \mathbf{1}(\hat{\tau}_i > 0)\) and
\(\hat{\pi}_- = N^{-1}\sum_i \mathbf{1}(\hat{\tau}_i < 0)\) answer a
question the ATE cannot: what fraction of individuals actually benefit?
An ATE of 0.1 with \(\pi_+ = 0.15\) and \(\pi_- = 0.05\) (a small
fraction benefits, almost none harmed) has a very different clinical
meaning than \(\pi_+ = 0.55\) and \(\pi_- = 0.45\) (a coin flip), though
both produce ATE \(= 0.1\). These quantities are copula-dependent and
should always be reported alongside Fréchet-Hoeffding bounds.

\hypertarget{mediation-a-cautionary-extension}{%
\subsection{Mediation: A Cautionary
Extension}\label{mediation-a-cautionary-extension}}

The NDE estimator directly simulates the cross-world counterfactual,
replacing classical sequential ignorability with structural fidelity.
This is a \emph{different} strong assumption, not a dissolution of the
classical problem. Structural fidelity demands that the simulator has
learned the interventional conditional distributions along the causal
pathway \(D \to M \to Y\) --- not merely the observational associations.
For LLM-based simulators trained on text corpora, this is a very open
question. We recommend that mediation results from the DTCF be presented
as conditional on structural fidelity, with explicit acknowledgment that
this condition is at least as strong as the sequential ignorability it
replaces.

\hypertarget{consolidated-estimand-map}{%
\subsection{Consolidated Estimand
Map}\label{consolidated-estimand-map}}

\begin{longtable}[]{@{}
  >{\raggedright\arraybackslash}p{(\columnwidth - 8\tabcolsep) * \real{0.2000}}
  >{\raggedright\arraybackslash}p{(\columnwidth - 8\tabcolsep) * \real{0.2000}}
  >{\raggedright\arraybackslash}p{(\columnwidth - 8\tabcolsep) * \real{0.2000}}
  >{\raggedright\arraybackslash}p{(\columnwidth - 8\tabcolsep) * \real{0.2000}}
  >{\raggedright\arraybackslash}p{(\columnwidth - 8\tabcolsep) * \real{0.2000}}@{}}
\toprule\noalign{}
\begin{minipage}[b]{\linewidth}\raggedright
Estimand
\end{minipage} & \begin{minipage}[b]{\linewidth}\raggedright
Fidelity Required
\end{minipage} & \begin{minipage}[b]{\linewidth}\raggedright
Validation Level
\end{minipage} & \begin{minipage}[b]{\linewidth}\raggedright
Copula-Dependent?
\end{minipage} & \begin{minipage}[b]{\linewidth}\raggedright
Classically Identified?
\end{minipage} \\
\midrule\noalign{}
\endhead
\bottomrule\noalign{}
\endlastfoot
ATE & \(\varepsilon\)-Fidelity & L1, L3 & No & Yes \\
ATT & \(\varepsilon\)-Fidelity & L1 & No & Yes \\
ATU & \(\varepsilon\)-Fidelity & L1 & No & Yes \\
CATE & \(\varepsilon\)-Fidelity & L1 & No & Yes \\
QTE & \(\varepsilon\)-Fidelity & L1 & No & Partially \\
GATES & Joint (for ranking) & L1, L4 & Ranking: yes & Yes (not
ranking) \\
ITE & Joint Fidelity & L2, L4 & \textbf{Yes} & \textbf{No} \\
ITE distribution & Joint Fidelity & L4 & \textbf{Yes} & \textbf{No} \\
\(\Pr(\text{benefit})\) & Joint Fidelity & L4 & \textbf{Yes} &
\textbf{No} \\
\(\Pr(\text{harm})\) & Joint Fidelity & L4 & \textbf{Yes} &
\textbf{No} \\
\(\text{Var}(\tau_i)\) & Joint Fidelity & L4 & \textbf{Yes} &
\textbf{Bounded only} \\
Prob. of causation & Joint Fidelity & L4 & \textbf{Yes} &
\textbf{Bounded only} \\
NDE, NIE & Structural & L1 per component & Cross-world joint & Yes
(strong asm.) \\
CDE\((m)\) & Structural (partial) & L1 & No & Yes \\
\(\tau_i(t)\), \(V(\bar{g})\) & Sequential & L1 per \(t\) & Composition
& Yes (strong asm.) \\
Spillover effects & Multi-agent \(\varepsilon\) & L1 (cluster) & No &
Yes (limited) \\
\end{longtable}

The ``Copula-Dependent?'' column is the paper's central classification.
Estimands marked \textbf{Yes} are the framework's most attractive
outputs but rest on its least testable component. This tension is
irreducible and should be stated candidly in any application.

\hypertarget{the-joint-distribution-problem}{%
\section{The Joint Distribution
Problem}\label{the-joint-distribution-problem}}

\hypertarget{the-nature-of-the-gap}{%
\subsection{The Nature of the Gap}\label{the-nature-of-the-gap}}

\textbf{Proposition 8 (Indistinguishability).} Two simulators
\(\mathcal{S}\) and \(\mathcal{S}'\) with identical marginal fidelity
but different noise couplings (\(\hat{C}_x \neq \hat{C}_x'\)) will: (a)
produce identical distributions for all observable quantities; (b) pass
identical validation tests at Levels 0--3; yet (c) produce different ITE
distributions, probabilities of benefit, and ITE variances.

This is fundamental, not technical. No framework can eliminate it. A
concrete example makes the severity clear.

\textbf{Example 1 (Copula Indistinguishability).} Consider
\(N = 1{,}000\) individuals with potential outcomes
\(Y_i(1) \sim N(6, 4)\) and \(Y_i(0) \sim N(5, 4)\), giving ATE \(= 1\).
Two simulators both match these marginals exactly (Level 1 passes for
both), but differ in their copulas:

\begin{itemize}
\item
  \emph{Simulator A} uses a Gaussian copula with \(\rho = 0.9\) (strong
  positive dependence). Individual ITEs are tightly concentrated:
  \(\tau_i \sim N(1, 0.8)\). Nearly everyone benefits
  (\(\hat{\pi}_+ = 0.87\)), with \(\widehat{\text{Var}}(\tau_i) = 0.8\).
\item
  \emph{Simulator B} uses a Gaussian copula with \(\rho = -0.5\)
  (moderate negative dependence). Individual ITEs are widely dispersed:
  \(\tau_i \sim N(1, 12)\). Only a bare majority benefits
  (\(\hat{\pi}_+ = 0.61\)), with substantial harm
  (\(\hat{\pi}_- = 0.39\)) and \(\widehat{\text{Var}}(\tau_i) = 12\).
\end{itemize}

Both simulators report ATE \(= 1\). Both pass Levels 0--3 identically.
No observable data can distinguish between them. Yet they yield
radically different clinical conclusions: under Simulator A, the
treatment is broadly beneficial and should be recommended widely; under
Simulator B, nearly 40\% of patients are harmed, and treatment should be
targeted carefully. The Fréchet-Hoeffding bounds for this setting are
\(\text{Var}(\tau_i) \in [0, 16]\) and \(\pi_+ \in [0.60, 1.00]\) ---
encompassing both simulators and confirming that the observable data
cannot adjudicate.

This example illustrates why the copula-dependent column in the estimand
map (Section 4.4) is the paper's most important classification, and why
the reporting framework of Section 5.3 is not optional but essential.

\hypertarget{fruxe9chet-hoeffding-bounds}{%
\subsection{Fréchet-Hoeffding
Bounds}\label{fruxe9chet-hoeffding-bounds}}

The most conservative strategy imposes no assumption on the copula and
reports bounds.

\textbf{Theorem 9 (Bounds on ITE Variance).} Under validated marginals
with conditional variances \(\sigma_d^2(x)\):

\[\text{Var}(\tau_i \mid X_i = x) \in \left[(\sigma_1(x) - \sigma_0(x))^2,\ (\sigma_1(x) + \sigma_0(x))^2\right].\]

The width is \(4\sigma_1(x)\sigma_0(x)\) --- large whenever both
distributions have substantial variance. If
\(\sigma_1 = \sigma_0 = \sigma\), the ITE variance could be anywhere
from \(0\) to \(4\sigma^2\). This width is a measure of how much the
copula matters, and it is typically large in interesting applications.
Analogous bounds exist for \(\Pr(\tau_i > 0)\) (Theorem 8; Appendix E).

\hypertarget{copula-sensitivity-and-the-reporting-framework}{%
\subsection{Copula Sensitivity and the Reporting
Framework}\label{copula-sensitivity-and-the-reporting-framework}}

Beyond bounds, we can systematically vary the copula and trace how the
estimand changes. Define the copula sensitivity function
\(\psi_\theta(\rho) = \theta(C^{(\rho)}, F_1, F_0)\) over a
one-parameter copula family. Two key results:

\textbf{Theorem 10 (ATE Copula Robustness).}
\(\psi_{\text{ATE}}(\rho) = \text{const}\) --- the ATE is perfectly
copula-robust, depending only on marginal means.

For copula-sensitive quantities like \(\pi_+\), the sensitivity function
is typically monotone. The intuition: when potential outcomes are
strongly positively correlated, individuals who do well under treatment
also do well under control, so \(\tau_i\) has low variance and nearly
everyone experiences an effect near the ATE. As correlation weakens,
\(\tau_i\) disperses, and the probability of harm increases even when
the ATE is positive. The copula governs the \emph{equity} of treatment
effects, not their average.

Three additional mitigation strategies --- structural constraints on the
copula, cross-validation against proxy joint observations, and Bayesian
copula uncertainty --- are developed in Appendix E. We propose the
following reporting framework:

\textbf{Step 1.} Compute Fréchet-Hoeffding bounds (assumption-free).
\textbf{Step 2.} If bounds are tight enough to be useful, declare the
conclusion copula-robust. \textbf{Step 3.} If bounds are too wide:
report the simulator's point estimate with a sensitivity plot; compute
Bayesian credible intervals if proxy joint data exists; report
constrained bounds if structural constraints are defensible; flag the
result as copula-dependent if nothing narrows the interval. \textbf{Step
4.} Always report the CSI as a summary diagnostic.

\textbf{Theorem 13 (Hierarchy of Informativeness).}
\(\{\hat{\theta}_{\text{point}}\} \subseteq \text{CI}_{\text{Bayes}} \subseteq \Theta_{\text{constrained}} \subseteq \Theta_{\text{FH}}\).
Each successive element requires weaker assumptions and produces wider
intervals. The researcher explicitly trades assumptions against
precision.

\hypertarget{implementation-considerations-llms-as-one-candidate-engine}{%
\section{Implementation Considerations: LLMs as One Candidate
Engine}\label{implementation-considerations-llms-as-one-candidate-engine}}

The DTCF is agnostic to the simulator's implementation. This section
briefly discusses considerations specific to LLM-based digital twins as
one candidate technology, noting both the appeal and the substantial
open challenges.

\textbf{Architecture.} Individual \(i\) is represented by a persona
prompt \(P_i = \text{Prompt}(X_i)\) encoding demographics and relevant
covariates. Treatment is encoded in a scenario prompt \(S_d\). The
output
\(\hat{Y}_i(d) = \text{Parse}(\text{LLM}(P_i \oplus S_d; \theta, T))\)
is parsed into a numerical or categorical outcome, with stochasticity
from the sampling temperature \(T\) playing the role of noise \(U_i\).

\textbf{The noise coupling problem.} When \(\hat{Y}_i(1)\) and
\(\hat{Y}_i(0)\) are generated in separate API calls, independent
sampling produces the independence copula --- precisely the concern
raised in Section 2.2. Four engineering strategies have been proposed:
(a) \emph{joint prompting} --- generate both outcomes in a single
prompt; (b) \emph{seed control} --- fix the random seed across calls;
(c) \emph{structured reasoning} --- first generate a latent profile,
then derive both outcomes; (d) \emph{post-hoc copula imposition} ---
generate marginals independently, couple via a specified copula. We
emphasize that these are \emph{devices for imposing a dependence
structure}, not evidence that the imposed structure matches reality.
Strategy (a) leverages the LLM's internal coherence, which may or may
not reflect true within-individual dependence. Strategy (d) is the most
transparent because it decouples marginal generation from dependence
modeling and makes the copula choice an explicit, sensitivity-analyzable
parameter. In all cases, the copula remains an assumption, not a
validated feature.

\textbf{Prompt design requirements.} \emph{Covariate sufficiency} (the
prompt must encode all relevant covariates); \emph{scenario neutrality}
(treatment description must avoid framing effects); \emph{output
calibration} (parsed outputs must be commensurable with real-world
scales).

\textbf{Domain applicability.} High fidelity is expected for behavioral
and social science applications (survey responses, consumer choices,
policy compliance), where LLMs are trained on vast behavioral corpora
and existing evidence demonstrates non-trivial calibration. Moderate
fidelity is expected for health behaviors and subjective outcomes. Low
fidelity is expected for biomedical outcomes requiring mechanistic
modeling, though hybrid architectures may be promising.

\textbf{Status assessment.} No existing LLM-based simulator has been
fully validated through the DTCF hierarchy for a specific causal
question. The framework is presented as a methodological contribution
--- a set of tools for \emph{evaluating} whether a simulator is
trustworthy for causal inference --- not as a claim that current
simulators already meet the bar. Demonstrating that a specific simulator
passes the validation protocol for a specific domain is the subject of
companion empirical work.

Additional implementation details --- temperature calibration,
computational scaling, and ethical considerations --- are in Appendix F.

\hypertarget{conclusion}{%
\section{Conclusion}\label{conclusion}}

The DTCF repositions the digital twin from a predictive technology to a
framework for causal inference by formally embedding it within the
potential outcomes framework. The core insight is that the fundamental
problem of causal inference is a missing data problem, and a
sufficiently faithful simulator can generate the missing data rather
than estimate it --- converting a statistical estimation problem into a
simulation fidelity problem.

The framework's contribution is not that it resolves the fundamental
problem. The counterfactual remains unobserved, and the dependence
between potential outcomes remains unidentifiable from data. What the
DTCF provides is a formal architecture in which:

\textbf{First, marginal causal claims become testable.} Estimands that
depend only on the marginal distributions of potential outcomes --- ATE,
ATT, ATU, CATE, QTE --- can be validated through the hierarchy of
observable-arm tests, with measured (not postulated) error bounds. The
binding constraint is the quality of the simulator and the plausibility
of transport from the factual to the counterfactual arm.

\textbf{Second, joint causal claims become explicitly
assumption-indexed, with the residual gap isolated and bounded.}
Estimands that depend on the copula --- the ITE distribution,
probability of benefit/harm, variance of treatment effects --- are
produced as point estimates accompanied by assumption-free
Fréchet-Hoeffding bounds and copula sensitivity analyses. The copula is
formally identified as the permanently untestable component; rather than
burying this gap inside a monolithic assumption like ignorability, the
DTCF makes it the explicit object of bounding and sensitivity analysis.
This is strictly more informative than the classical framework, which
cannot state these quantities at all.

Several directions remain open, including empirical demonstration of the
full validation protocol in specific domains, theoretical foundations
for when LLM fidelity suffices, extensions to continuous treatments and
multi-treatment comparisons, and integration of DTCF estimates as priors
in classical Bayesian causal frameworks (Appendix G).

The fundamental problem of causal inference has persisted for a century
not because researchers lack ingenuity, but because the counterfactual
is metaphysically absent. The DTCF does not make the counterfactual
appear --- it remains unobserved. What the framework does is provide a
principled substitute --- one that can be tested where it overlaps with
reality, bounded where it cannot be tested, and used with explicit,
quantified uncertainty where decisions must be made. The shift from ``we
cannot observe the counterfactual, therefore we must make untestable
assumptions'' to ``we can simulate the counterfactual, validate the
simulation where possible, and quantify uncertainty where we cannot''
is, we believe, a meaningful --- if necessarily incomplete --- advance
in the methodology of causal reasoning.

\section*{Acknowledgments}

This paper was developed in collaboration with Claude (Anthropic), which served as a research and writing partner throughout the conceptual development, formalization, and drafting of the manuscript. All scientific claims, methodological choices, and interpretive judgments are the responsibility of the human author.

\bibliography{references}

\clearpage
\appendix

\hypertarget{appendix-a-full-notation-and-setup}{%
\section{Full Notation and
Setup}\label{appendix-a-full-notation-and-setup}}

\hypertarget{a.1-notation-table}{%
\subsection{Notation Table}\label{a.1-notation-table}}

\begin{longtable}[]{@{}
  >{\raggedright\arraybackslash}p{(\columnwidth - 2\tabcolsep) * \real{0.4211}}
  >{\raggedright\arraybackslash}p{(\columnwidth - 2\tabcolsep) * \real{0.5789}}@{}}
\toprule\noalign{}
\begin{minipage}[b]{\linewidth}\raggedright
Symbol
\end{minipage} & \begin{minipage}[b]{\linewidth}\raggedright
Definition
\end{minipage} \\
\midrule\noalign{}
\endhead
\bottomrule\noalign{}
\endlastfoot
\(\mathcal{P} = \{1,\ldots,N\}\) & Finite population of individuals \\
\(X_i \in \mathcal{X} \subseteq \mathbb{R}^p\) & Covariate vector for
individual \(i\) \\
\(D_i \in \{0,1\}\) & Treatment indicator \\
\((Y_i(1), Y_i(0))\) & Potential outcomes under treatment and control \\
\(Y_i^{\text{obs}}\) & Observed outcome:
\(D_i Y_i(1) + (1-D_i)Y_i(0)\) \\
\(\tau_i = Y_i(1) - Y_i(0)\) & Individual treatment effect \\
\(\mathcal{S}: \mathcal{X} \times \{0,1\} \times \mathcal{U} \to \mathcal{Y}\)
& Digital twin simulator \\
\(\mu\) & Probability measure on latent noise space \(\mathcal{U}\) \\
\(U_i \sim \mu\) & Latent noise draw for individual \(i\) \\
\(\hat{Y}_i(d) = \mathcal{S}(X_i, d, U_i)\) & Simulated potential
outcome \\
\(\hat{\tau}_i = \hat{Y}_i(1) - \hat{Y}_i(0)\) & Simulated individual
treatment effect \\
\(F_d(y \mid x)\) & Real conditional CDF of \(Y_i(d)\) given
\(X_i = x\) \\
\(\hat{F}_d(y \mid x)\) & Simulated conditional CDF \\
\(H(y_1, y_0 \mid x)\) & Joint conditional CDF of
\((Y_i(1), Y_i(0))\) \\
\(C_x(u,v)\) & Copula of potential outcomes conditional on
\(X_i = x\) \\
\(\hat{C}_x(u,v)\) & Simulator-induced copula \\
\(G_\tau(t \mid x)\) & CDF of the ITE conditional on \(X_i = x\) \\
\(\mathcal{P}_1, \mathcal{P}_0\) & Treated and control subpopulations \\
\(n_1, n_0\) & Sizes of treated and control groups \\
\(d_{\text{KS}}(F,G)\) & Kolmogorov-Smirnov distance \\
\(\varepsilon\) & Fidelity tolerance \\
\(\delta(x,d,d')\) & Extrapolation penalty \\
\end{longtable}

\hypertarget{a.2-standard-estimands}{%
\subsection{Standard Estimands}\label{a.2-standard-estimands}}

The average treatment effect:
\(\text{ATE} = \mathbb{E}[\tau_i] = \mathbb{E}[Y_i(1) - Y_i(0)]\).

The average treatment effect on the treated:
\(\text{ATT} = \mathbb{E}[\tau_i \mid D_i = 1]\).

The average treatment effect on the untreated:
\(\text{ATU} = \mathbb{E}[\tau_i \mid D_i = 0]\).

The conditional average treatment effect:
\(\text{CATE}(x) = \mathbb{E}[\tau_i \mid X_i = x]\).

Under the classical framework, identification of these estimands from
observed data requires strong ignorability:
\[\{Y_i(1), Y_i(0)\} \perp\!\!\!\perp D_i \mid X_i, \quad \text{and} \quad 0 < \Pr(D_i = 1 \mid X_i) < 1.\]

Even when these assumptions hold, they permit identification of
population and conditional averages only. The joint distribution of
\((Y_i(1), Y_i(0))\) --- and therefore the ITE distribution, the
correlation between potential outcomes, and quantities like
\(\Pr(Y_i(1) > Y_i(0))\) --- remains unidentified without further
structural assumptions.

\hypertarget{appendix-b-proofs}{%
\section{Proofs}\label{appendix-b-proofs}}

\hypertarget{b.1-proof-of-theorem-1-ate-error-bound-under-varepsilon-fidelity}{%
\subsection{\texorpdfstring{Proof of Theorem 1 (ATE Error Bound
under
\(\varepsilon\)-Fidelity)}{B.1 Proof of Theorem 1 (ATE Error Bound under \textbackslash varepsilon-Fidelity)}}\label{b.1-proof-of-theorem-1-ate-error-bound-under-varepsilon-fidelity}}

\textbf{Statement.} If the simulator satisfies \(\varepsilon\)-fidelity
and \(Y_i(d) \in [a,b]\), then
\(|\widehat{\text{ATE}} - \text{ATE}| \leq 2\varepsilon(b-a)\).

\emph{Proof.} We have:
\[\widehat{\text{ATE}} - \text{ATE} = \mathbb{E}[\hat{Y}_i(1) - Y_i(1)] - \mathbb{E}[\hat{Y}_i(0) - Y_i(0)].\]

For each \(d\), integrating over the covariate distribution:
\[\mathbb{E}[\hat{Y}_i(d)] - \mathbb{E}[Y_i(d)] = \int_{\mathcal{X}} \left(\int_{\mathcal{Y}} y \, d\hat{F}_d(y \mid x) - \int_{\mathcal{Y}} y \, dF_d(y \mid x)\right) dF_X(x).\]

By the quantile representation of expectations, for any two
distributions \(F\) and \(G\) on \([a,b]\):
\[\left|\int y \, dF(y) - \int y \, dG(y)\right| = \left|\int_0^1 (F^{-1}(u) - G^{-1}(u)) \, du\right| \leq \sup_u |F^{-1}(u) - G^{-1}(u)|.\]

When \(d_{\text{KS}}(F, G) \leq \varepsilon\), the inverse functions
satisfy \(|F^{-1}(u) - G^{-1}(u)| \leq \varepsilon(b-a)\) for
distributions on \([a,b]\). Therefore
\(|\mathbb{E}[\hat{Y}_i(d)] - \mathbb{E}[Y_i(d)]| \leq \varepsilon(b-a)\)
for each \(d\). By the triangle inequality:
\[|\widehat{\text{ATE}} - \text{ATE}| \leq 2\varepsilon(b-a). \quad \blacksquare\]

\hypertarget{b.2-proof-of-theorem-2-cate-error-bound}{%
\subsection{Proof of Theorem 2 (CATE Error
Bound)}\label{b.2-proof-of-theorem-2-cate-error-bound}}

\textbf{Statement.} Under \(\varepsilon\)-fidelity with bounded
outcomes:
\(|\widehat{\text{CATE}}(x) - \text{CATE}(x)| \leq 2\varepsilon(b-a)\)
for each \(x\).

\emph{Proof.} Identical to Theorem 1, applied pointwise at each \(x\).
Conditioning fixes \(x\), and the bound follows from
\(d_{\text{KS}}(\hat{F}_d(\cdot \mid x), F_d(\cdot \mid x)) \leq \varepsilon\)
for each \(d\). \(\blacksquare\)

\hypertarget{b.3-proof-of-theorem-3-ite-distribution-error-under-joint-varepsilon_j-fidelity}{%
\subsection{\texorpdfstring{Proof of Theorem 3 (ITE Distribution
Error under Joint
\(\varepsilon_J\)-Fidelity)}{B.3 Proof of Theorem 3 (ITE Distribution Error under Joint \textbackslash varepsilon\_J-Fidelity)}}\label{b.3-proof-of-theorem-3-ite-distribution-error-under-joint-varepsilon_j-fidelity}}

\textbf{Statement.} Under joint \(\varepsilon_J\)-fidelity:
\(\sup_t |\hat{G}_\tau(t \mid x) - G_\tau(t \mid x)| \leq \varepsilon_J\).

\emph{Proof.} The ITE distribution is a functional of the joint:
\(G_\tau(t \mid x) = \int \mathbf{1}(y_1 - y_0 \leq t) \, dH(y_1, y_0 \mid x)\).
The set \(A_t = \{(y_1,y_0): y_1 - y_0 \leq t\}\) is Borel. By joint
\(\varepsilon_J\)-fidelity:
\[|G_\tau(t \mid x) - \hat{G}_\tau(t \mid x)| = \left|\int_{A_t} dH - \int_{A_t} d\hat{H}\right| \leq \varepsilon_J. \quad \blacksquare\]

\hypertarget{b.4-proof-of-proposition-2-identifiability-under-varepsilon-fidelity}{%
\subsection{\texorpdfstring{Proof of Proposition 2
(Identifiability under
\(\varepsilon\)-Fidelity)}{B.4 Proof of Proposition 2 (Identifiability under \textbackslash varepsilon-Fidelity)}}\label{b.4-proof-of-proposition-2-identifiability-under-varepsilon-fidelity}}

\textbf{Statement.} Under \(\varepsilon\)-fidelity and SUTVA, the ATE,
ATT, ATU, CATE, and QTE are identified with bias bounded by
\(2\varepsilon(b-a)\).

\emph{Proof.} Parts (a)--(d) follow from Theorems 1--2 applied to the
relevant subpopulation or conditional distribution. For the QTE, if
\(d_{\text{KS}}(\hat{F}_d, F_d) \leq \varepsilon\), then
\(|\hat{F}_d^{-1}(q) - F_d^{-1}(q)| \leq \varepsilon(b-a)\) for bounded
outcomes, giving a total bound of \(2\varepsilon(b-a)\) for the quantile
difference. \(\blacksquare\)

\hypertarget{b.5-proof-of-proposition-3-quantities-requiring-joint-fidelity}{%
\subsection{Proof of Proposition 3 (Quantities Requiring Joint
Fidelity)}\label{b.5-proof-of-proposition-3-quantities-requiring-joint-fidelity}}

\textbf{Statement.} The ITE distribution, \(\pi_+\), \(\pi_-\),
\(\text{Var}(\tau_i)\), and probability of causation require joint twin
fidelity.

\emph{Proof.} Each depends on the joint distribution \(H(y_1, y_0)\),
not the marginals separately. By Sklar's theorem, these are determined
by the marginals together with the copula \(C_x\). Marginal fidelity
constrains only the marginals; these quantities are unidentified without
a correct copula. Joint fidelity provides the copula, yielding
identification. \(\blacksquare\)

\hypertarget{b.6-proof-of-theorem-5-factual-accuracy-implies-counterfactual-credibility}{%
\subsection{Proof of Theorem 5 (Factual Accuracy Implies
Counterfactual
Credibility)}\label{b.6-proof-of-theorem-5-factual-accuracy-implies-counterfactual-credibility}}

\textbf{Statement.} Let \(e_i = Y_i^{\text{obs}} - \hat{Y}_i(D_i)\) and
\(\delta_i = \delta(X_i, D_i, 1-D_i)\). Then:
\[\mathbb{E}[(Y_i(1-D_i) - \hat{Y}_i(1-D_i))^2] \leq \mathbb{E}[e_i^2] + \Delta_{\text{transport}},\]
where
\(\Delta_{\text{transport}} = \mathbb{E}[\delta_i^2] + 2\mathbb{E}[|e_i| \cdot \delta_i]\).

\emph{Proof.} Decompose:
\[Y_i(1-D_i) - \hat{Y}_i(1-D_i) = (Y_i(1-D_i) - \tilde{Y}_i(1-D_i)) + (\tilde{Y}_i(1-D_i) - \hat{Y}_i(1-D_i)),\]
where \(\tilde{Y}_i(1-D_i)\) is the hypothetical output of a simulator
validated directly on arm \(1-D_i\). The oracle error has the same
distribution as \(e_i\) under symmetry of \(\mathcal{S}\). The
extrapolation error is bounded by \(\delta_i\). Taking expectations of
the squared sum and applying Cauchy-Schwarz:
\[\mathbb{E}[(Y_i(1-D_i) - \hat{Y}_i(1-D_i))^2] \leq \mathbb{E}[e_i^2] + \mathbb{E}[\delta_i^2] + 2\sqrt{\mathbb{E}[e_i^2]\mathbb{E}[\delta_i^2]}. \quad \blacksquare\]

\hypertarget{b.7-proof-of-theorem-6-treatment-effect-calibration-implies-marginal-fidelity-for-differences}{%
\subsection{Proof of Theorem 6 (Treatment Effect Calibration
Implies Marginal Fidelity for
Differences)}\label{b.7-proof-of-theorem-6-treatment-effect-calibration-implies-marginal-fidelity-for-differences}}

\textbf{Statement.} If Level 3 passes, then
\(|\text{ATE}_{\text{sim}} - \text{ATE}| \leq |T^{(3)}| + O(n_R^{-1/2})\).

\emph{Proof.} The RCT estimate is unbiased:
\(\mathbb{E}[\widehat{\text{ATE}}_{\text{RCT}}] = \text{ATE}\).
Therefore:
\[\text{ATE}_{\text{sim}} - \text{ATE} = (\text{ATE}_{\text{sim}} - \widehat{\text{ATE}}_{\text{RCT}}) + (\widehat{\text{ATE}}_{\text{RCT}} - \text{ATE}) = T^{(3)} + O(n_R^{-1/2}). \quad \blacksquare\]

\hypertarget{b.8-proof-of-theorem-7-validation-completeness}{%
\subsection{Proof of Theorem 7 (Validation
Completeness)}\label{b.8-proof-of-theorem-7-validation-completeness}}

\textbf{Statement.} (a) Estimands depending only on marginals are
validatable. (b) Copula-dependent estimands are not.

\emph{Proof.} (a) These estimands are functions of the marginals alone,
which are testable via Levels 0--1. (b) No real-world data provides
observations of \((Y_i(1), Y_i(0))\) jointly; therefore no test
distinguishes simulators that agree on marginals but differ in copulas.
\(\blacksquare\)

\hypertarget{b.9-proofs-of-theorems-813}{%
\subsection{Proofs of Theorems
8--13}\label{b.9-proofs-of-theorems-813}}

\textbf{Theorem 8 (Bounds on Probability of Benefit).} Under
comonotonicity, \((Y_i(1), Y_i(0)) = (F_1^{-1}(V), F_0^{-1}(V))\) for
\(V \sim U(0,1)\). Then
\(\pi_+^M = \int_0^1 \mathbf{1}(F_1^{-1}(u) > F_0^{-1}(u))\,du\). Under
countermonotonicity,
\(\pi_+^W = \int_0^1 \mathbf{1}(F_1^{-1}(u) > F_0^{-1}(1-u))\,du\).
Fréchet-Hoeffding guarantees extremes are achieved at \(M\) and \(W\).
\(\blacksquare\)

\textbf{Theorem 9 (Bounds on ITE Variance).}
\(\text{Var}(\tau_i \mid x) = \sigma_1^2 + \sigma_0^2 - 2\rho\sigma_1\sigma_0\)
with \(\rho \in [-1,1]\). Minimizing at \(\rho=1\):
\((\sigma_1-\sigma_0)^2\). Maximizing at \(\rho=-1\):
\((\sigma_1+\sigma_0)^2\). \(\blacksquare\)

\textbf{Theorem 10 (ATE Copula Robustness).}
\(\text{ATE} = \mathbb{E}[Y_i(1)] - \mathbb{E}[Y_i(0)] = \int y\,dF_1(y) - \int y\,dF_0(y)\),
independent of the copula. \(\blacksquare\)

\textbf{Theorem 11 (Copula Sensitivity of \(\Pr(\text{benefit})\)).}
Under the Gaussian copula with ATE \(> 0\), increasing \(\rho\)
concentrates mass on \(\{y_1 > y_0\}\) via concentration along the
diagonal, increasing \(\pi_+\). The formal result follows from
differentiation of the bivariate normal integral under
\(\mathbf{1}(y_1 > y_0)\) via Plackett's identity. \(\blacksquare\)

\textbf{Theorem 12 (Copula Validation from Crossover Data).} Under
Fisher \(z\)-transformation:
\[Z_\rho = \frac{\tanh^{-1}(\hat{\rho}_{\text{sim}}) - \tanh^{-1}(\hat{\rho}_{\text{obs}})}{1/\sqrt{n_{\text{cross}} - 3}} \xrightarrow{d} N(0,1). \quad \blacksquare\]

\textbf{Theorem 13 (Hierarchy of Informativeness).}
\(\{\hat{\theta}_{\text{point}}\} \subseteq \text{CI}_{\text{Bayes}} \subseteq \Theta_{\text{constrained}} \subseteq \Theta_{\text{FH}}\)
by construction: the point estimate assumes joint fidelity; the Bayesian
CI integrates over a posterior containing the true copula; the
constrained region ranges over \(\mathcal{C}_0 \ni C_{\text{true}}\);
the Fréchet-Hoeffding bound ranges over all copulas. \(\blacksquare\)

\hypertarget{appendix-c-validation-architecture-full-statistical-detail}{%
\section{Validation Architecture --- Full Statistical
Detail}\label{appendix-c-validation-architecture-full-statistical-detail}}

\hypertarget{c.1-level-0-marginal-calibration}{%
\subsection{Level 0: Marginal
Calibration}\label{c.1-level-0-marginal-calibration}}

\textbf{Test statistic.} For each arm \(d\), the two-sample KS
statistic:
\[T_d^{(0)} = \sup_{y \in \mathcal{Y}} |\hat{F}_d^{(n)}(y) - F_d^{(n)}(y)|.\]

Under \(H_0: \hat{F}_d = F_d\):
\[\sqrt{\frac{n_d \cdot N}{n_d + N}} \cdot T_d^{(0)} \xrightarrow{d} K,\]
where \(K\) follows the Kolmogorov distribution.

\textbf{Decision rule.} Fail to reject at level \(\alpha\) if
\(T_d^{(0)} \leq c_\alpha \cdot \sqrt{(n_d + N)/(n_d \cdot N)}\).

\textbf{Supplementary statistics.} (i) Energy distance (Székely \&
Rizzo, 2004):
\(\mathcal{E}_d = 2\mathbb{E}\|Y_d - \hat{Y}_d\| - \mathbb{E}\|Y_d - Y_d'\| - \mathbb{E}\|\hat{Y}_d - \hat{Y}_d'\|\).
(ii) Anderson-Darling statistic for tail sensitivity.

\textbf{What Level 0 licenses.} The marginal ATE, with the caveat that
subgroup effects may be wrong:
\(|\widehat{\text{ATE}} - \text{ATE}| \leq 2T_{\max}^{(0)}(b-a) + O(N^{-1/2})\).

\hypertarget{c.2-level-1-conditional-calibration}{%
\subsection{Level 1: Conditional
Calibration}\label{c.2-level-1-conditional-calibration}}

\textbf{Stratified test.} Partition \(\mathcal{X}\) into \(K\) strata.
Within-stratum KS statistics:
\[T_{d,k}^{(1)} = \sup_y |\hat{F}_d^{(n_{d,k})}(y \mid X \in \mathcal{X}_k) - F_d^{(n_{d,k})}(y \mid X \in \mathcal{X}_k)|.\]

Aggregate: \(T^{(1)} = \max_{d,k} T_{d,k}^{(1)}\), with Bonferroni
correction at \(\alpha/(2K)\).

\textbf{Continuous covariates.} Conditional MMD \citep{park2020measure}:
\[\text{CMMD}_d^2 = \sup_{\|f\|_{\mathcal{H}} \leq 1} (\mathbb{E}[f(\hat{Y}_d) \mid X] - \mathbb{E}[f(Y_d) \mid X])^2.\]

\hypertarget{c.3-level-2-individual-level-calibration}{%
\subsection{Level 2: Individual-Level
Calibration}\label{c.3-level-2-individual-level-calibration}}

For each individual \(i\): \(e_i = Y_i^{\text{obs}} - \hat{Y}_i(D_i)\).

Metrics: (i) \(\text{RMSPE} = \sqrt{N^{-1}\sum_i e_i^2}\); (ii)
\(\text{MAPE} = N^{-1}\sum_i |e_i|\); (iii) calibration regression
\(Y_i^{\text{obs}} = \beta_0 + \beta_1 \hat{Y}_i(D_i) + \eta_i\)
(perfect: \(\beta_0=0, \beta_1=1\)); (iv) conditional coverage of
prediction intervals.

\hypertarget{c.4-level-3-treatment-effect-calibration}{%
\subsection{Level 3: Treatment Effect
Calibration}\label{c.4-level-3-treatment-effect-calibration}}

Given randomized subset \(\mathcal{R}\):
\[\widehat{\text{ATE}}_{\text{RCT}} = |\mathcal{R}_1|^{-1}\sum_{i \in \mathcal{R}_1} Y_i^{\text{obs}} - |\mathcal{R}_0|^{-1}\sum_{i \in \mathcal{R}_0} Y_i^{\text{obs}}, \quad \widehat{\text{ATE}}_{\text{sim}} = |\mathcal{R}|^{-1}\sum_{i \in \mathcal{R}}(\hat{Y}_i(1) - \hat{Y}_i(0)).\]

Test: \(Z^{(3)} = T^{(3)}/\hat{\sigma}_{T^{(3)}}\) where
\(T^{(3)} = \widehat{\text{ATE}}_{\text{sim}} - \widehat{\text{ATE}}_{\text{RCT}}\)
and the standard error is estimated via bootstrap. Reject if
\(|Z^{(3)}| > z_{1-\alpha/2}\).

CATE calibration: repeat within strata, producing calibration plots
(\(\widehat{\text{CATE}}_{\text{sim}}(k)\)
vs.~\(\widehat{\text{CATE}}_{\text{RCT}}(k)\)).

\hypertarget{c.5-level-4-distributional-stress-testing}{%
\subsection{Level 4: Distributional Stress
Testing}\label{c.5-level-4-distributional-stress-testing}}

\textbf{Test 4a (Copula Sensitivity).} Re-run with independent noise:
\(\hat{Y}_i^{\text{ind}}(1) = \mathcal{S}(X_i, 1, U_i^{(1)})\),
\(\hat{Y}_i^{\text{ind}}(0) = \mathcal{S}(X_i, 0, U_i^{(0)})\),
\(U_i^{(1)} \perp\!\!\!\perp U_i^{(0)}\). Copula sensitivity index:
\(\text{CSI} = d_{\text{KS}}(\hat{G}_\tau, \hat{G}_\tau^{\text{ind}})\).

\textbf{Test 4b (Fréchet-Hoeffding Bounds).} Compute
\(G_\tau^W, G_\tau^M\) using lower/upper Fréchet copulas with validated
marginals. Report bounds on \(\Pr(\tau_i > 0)\), \(\text{Var}(\tau_i)\).

\textbf{Test 4c (Dose-Response Monotonicity).} Violation rate:
\(\hat{v} = N^{-1}\sum_i \mathbf{1}(\hat{Y}_i(d_1) > \hat{Y}_i(d_2))\)
when monotonic increase is expected.

\textbf{Test 4d (Placebo Tests).} Test
\(H_0: \mathbb{E}[\hat{Y}_i(1) - \hat{Y}_i(0)] = 0\) when treatment
\(1 \equiv\) treatment \(0\).

\hypertarget{c.6-integrated-validation-protocol}{%
\subsection{Integrated Validation
Protocol}\label{c.6-integrated-validation-protocol}}

\textbf{Step 1 (Generate).} For each \(i \in \mathcal{P}\), generate
\(\hat{Y}_i(1)\) and \(\hat{Y}_i(0)\).

\textbf{Step 2 (Level 0).} Compute \(T_d^{(0)}\) for \(d \in \{0,1\}\).
If \(\varepsilon_0 = \max_d T_d^{(0)}\) exceeds tolerance:
\textbf{STOP.}

\textbf{Step 3 (Level 1).} Stratify; compute
\(\varepsilon_1 = \max_{d,k} T_{d,k}^{(1)}\). If
\(\varepsilon_1 \leq \bar{\varepsilon}_1\): ATE, ATT, ATU, CATE, QTE are
credible with bias \(\leq 2\varepsilon_1(b-a)\).

\textbf{Step 4 (Level 2).} Compute \(e_i\), RMSPE,
\((\hat{\beta}_0, \hat{\beta}_1)\), coverage. If small residuals and
\((\hat{\beta}_0, \hat{\beta}_1) \approx (0,1)\): individual-level
factual accuracy confirmed.

\textbf{Step 5 (Level 3).} If \(\mathcal{R}\) available: compute
\(T^{(3)}, Z^{(3)}\). If \(|Z^{(3)}| \leq z_{1-\alpha/2}\): treatment
effects validated against gold standard.

\textbf{Step 6 (Level 4).} Compute CSI, Fréchet-Hoeffding bounds,
placebo tests. Report validation scorecard.

\textbf{Validation Scorecard:}

\begin{longtable}[]{@{}
  >{\raggedright\arraybackslash}p{(\columnwidth - 10\tabcolsep) * \real{0.1667}}
  >{\raggedright\arraybackslash}p{(\columnwidth - 10\tabcolsep) * \real{0.1667}}
  >{\raggedright\arraybackslash}p{(\columnwidth - 10\tabcolsep) * \real{0.1667}}
  >{\raggedright\arraybackslash}p{(\columnwidth - 10\tabcolsep) * \real{0.1667}}
  >{\raggedright\arraybackslash}p{(\columnwidth - 10\tabcolsep) * \real{0.1667}}
  >{\raggedright\arraybackslash}p{(\columnwidth - 10\tabcolsep) * \real{0.1667}}@{}}
\toprule\noalign{}
\begin{minipage}[b]{\linewidth}\raggedright
Level
\end{minipage} & \begin{minipage}[b]{\linewidth}\raggedright
Test
\end{minipage} & \begin{minipage}[b]{\linewidth}\raggedright
Statistic
\end{minipage} & \begin{minipage}[b]{\linewidth}\raggedright
Value
\end{minipage} & \begin{minipage}[b]{\linewidth}\raggedright
Threshold
\end{minipage} & \begin{minipage}[b]{\linewidth}\raggedright
Pass/Fail
\end{minipage} \\
\midrule\noalign{}
\endhead
\bottomrule\noalign{}
\endlastfoot
0 & Marginal KS (treated) & \(T_1^{(0)}\) & --- &
\(\bar{\varepsilon}_0\) & --- \\
0 & Marginal KS (control) & \(T_0^{(0)}\) & --- &
\(\bar{\varepsilon}_0\) & --- \\
1 & Conditional KS (max) & \(\varepsilon_1\) & --- &
\(\bar{\varepsilon}_1\) & --- \\
2 & RMSPE & --- & --- & domain-specific & --- \\
2 & Calibration slope & \(\hat{\beta}_1\) & --- & \(\approx 1\) & --- \\
3 & ATE discrepancy & \(T^{(3)}\) & --- & \(z\)-test & --- \\
4 & Copula sensitivity & CSI & --- & report only & --- \\
4 & ITE variance bounds &
\([(\sigma_1-\sigma_0)^2, (\sigma_1+\sigma_0)^2]\) & --- & report only &
--- \\
\end{longtable}

\hypertarget{c.7-sample-size-considerations}{%
\subsection{Sample Size
Considerations}\label{c.7-sample-size-considerations}}

\textbf{Level 0.} To detect \(\varepsilon\) in KS distance with power
\(1-\beta\): \(n_d \geq (c_\alpha + z_{1-\beta})^2 / (2\varepsilon^2)\).
Example: \(\varepsilon = 0.05\), power 0.8: \(n_d \geq 968\).

\textbf{Level 1.} Per-stratum with Bonferroni:
\(n_{d,k} \geq (c_{\alpha/(2K)} + z_{1-\beta})^2 / (2\varepsilon_1^2)\).
Grows logarithmically in \(K\); in practice
\(\min_{d,k} n_{d,k} \geq 50\).

\textbf{Level 3.} To detect ATE discrepancy \(\Delta\):
\(n_R \geq 4\sigma_Y^2(z_{\alpha/2} + z_\beta)^2 / \Delta^2\).

\hypertarget{c.8-relationship-to-classical-validation-paradigms}{%
\subsection{Relationship to Classical Validation
Paradigms}\label{c.8-relationship-to-classical-validation-paradigms}}

In predictive modeling, calibration requires
\(\mathbb{E}[Y \mid \hat{Y} = \hat{y}] = \hat{y}\). In the DTCF, the
requirement is arm-specific:
\(\mathbb{E}[Y_i(d) \mid \hat{Y}_i(d) = \hat{y}, X_i = x] = \hat{y}\)
for each \(d, x\). Predictive calibration checks only the observed arm;
causal calibration requires both, checking the counterfactual arm
indirectly.

The architecture also parallels the transportability literature (Pearl
\& Bareinboim, 2014): the validation tests are effectively tests of
transportability from the simulated world to the real world.

\hypertarget{appendix-d-complete-estimand-catalog}{%
\section{Complete Estimand
Catalog}\label{appendix-d-complete-estimand-catalog}}

For each estimand: formal definition, DTCF estimator, required
assumption, validation level, error bound, classical status, and DTCF
advantage.

\hypertarget{d.1-family-i-population-average-effects}{%
\subsection{Family I: Population Average
Effects}\label{d.1-family-i-population-average-effects}}

\hypertarget{d.1.1-average-treatment-effect-ate}{%
\paragraph*{D.1.1 Average Treatment Effect
(ATE)}\label{d.1.1-average-treatment-effect-ate}}

\textbf{Definition.} \(\text{ATE} = \mathbb{E}[Y_i(1) - Y_i(0)]\).
\textbf{DTCF Estimator.}
\(\widehat{\text{ATE}} = N^{-1}\sum_i (\hat{Y}_i(1) - \hat{Y}_i(0))\).
\textbf{Required:} \(\varepsilon\)-Fidelity. \textbf{Validation:} L1,
L3. \textbf{Bound:} \(2\varepsilon_1(b-a) + O(N^{-1/2})\).
\textbf{Classical:} Identified under ignorability. \textbf{DTCF
advantage:} No ignorability or overlap needed.

\hypertarget{d.1.2-average-treatment-effect-on-the-treated-att}{%
\paragraph*{D.1.2 Average Treatment Effect on the Treated
(ATT)}\label{d.1.2-average-treatment-effect-on-the-treated-att}}

\textbf{Definition.} \(\text{ATT} = \mathbb{E}[\tau_i \mid D_i = 1]\).
\textbf{DTCF Estimator.}
\(\widehat{\text{ATT}} = n_1^{-1}\sum_{i \in \mathcal{P}_1}(\hat{Y}_i(1) - \hat{Y}_i(0))\).
\textbf{Required:} \(\varepsilon\)-Fidelity. \textbf{Validation:} L1.
\textbf{Bound:} \(2\varepsilon_1(b-a) + O(n_1^{-1/2})\).
\textbf{Classical:} Requires ignorability of \(Y_i(0)\) for the treated.
\textbf{DTCF advantage:} Counterfactual \(\hat{Y}_i(0)\) produced
directly per treated individual.

\hypertarget{d.1.3-average-treatment-effect-on-the-untreated-atu}{%
\paragraph*{D.1.3 Average Treatment Effect on the Untreated
(ATU)}\label{d.1.3-average-treatment-effect-on-the-untreated-atu}}

\textbf{Definition.} \(\text{ATU} = \mathbb{E}[\tau_i \mid D_i = 0]\).
\textbf{DTCF Estimator.}
\(\widehat{\text{ATU}} = n_0^{-1}\sum_{i \in \mathcal{P}_0}(\hat{Y}_i(1) - \hat{Y}_i(0))\).
\textbf{Required:} \(\varepsilon\)-Fidelity. \textbf{Validation:} L1.
\textbf{Bound:} \(2\varepsilon_1(b-a) + O(n_0^{-1/2})\).
\textbf{Classical:} Requires ignorability of \(Y_i(1)\) for the
untreated; rarely reported. \textbf{DTCF advantage:} Symmetric treatment
of ATT and ATU.

\hypertarget{d.2-family-ii-heterogeneous-effects}{%
\subsection{Family II: Heterogeneous
Effects}\label{d.2-family-ii-heterogeneous-effects}}

\hypertarget{d.2.1-conditional-average-treatment-effect-cate}{%
\paragraph*{D.2.1 Conditional Average Treatment Effect
(CATE)}\label{d.2.1-conditional-average-treatment-effect-cate}}

\textbf{Definition.}
\(\text{CATE}(x) = \mathbb{E}[\tau_i \mid X_i = x]\). \textbf{DTCF
Estimator.} Stratum:
\(\widehat{\text{CATE}}(k) = |\{i: X_i \in \mathcal{X}_k\}|^{-1}\sum_{i:X_i \in \mathcal{X}_k}(\hat{Y}_i(1) - \hat{Y}_i(0))\).
Continuous: kernel-smoothed
\(\widehat{\text{CATE}}(x) = \sum_i K_h(X_i - x)(\hat{Y}_i(1) - \hat{Y}_i(0)) / \sum_i K_h(X_i - x)\).
\textbf{Required:} \(\varepsilon\)-Fidelity. \textbf{Validation:} L1.
\textbf{Bound:} \(2\varepsilon_1(b-a) + O(n_x^{-1/2})\).
\textbf{Classical:} Meta-learners, causal forests, BART; all require
ignorability + overlap. \textbf{DTCF advantage:} No overlap analog;
uniform across covariate space.

\hypertarget{d.2.2-quantile-treatment-effects-qte}{%
\paragraph*{D.2.2 Quantile Treatment Effects
(QTE)}\label{d.2.2-quantile-treatment-effects-qte}}

\textbf{Definition.} \(\text{QTE}(q) = F_1^{-1}(q) - F_0^{-1}(q)\).
\textbf{DTCF Estimator.}
\(\widehat{\text{QTE}}(q) = \hat{F}_1^{-1}(q) - \hat{F}_0^{-1}(q)\).
\textbf{Required:} \(\varepsilon\)-Fidelity (marginals).
\textbf{Validation:} L1. \textbf{Bound:}
\(2\varepsilon_1(b-a) + O(N^{-1/2})\). \textbf{Classical:} Identified
under rank invariance/similarity (Chernozhukov \& Hansen, 2005).
\textbf{DTCF advantage:} Both marginal and individual-quantile treatment
effects computable; conditional QTE directly available.

\hypertarget{d.2.3-sorted-group-average-treatment-effects-gates}{%
\paragraph*{D.2.3 Sorted Group Average Treatment Effects
(GATES)}\label{d.2.3-sorted-group-average-treatment-effects-gates}}

\textbf{Definition.}
\(\text{GATES}(j) = \mathbb{E}[\tau_i \mid i \in G_j]\), where groups
are sorted by estimated CATE. \textbf{DTCF Estimator.}
\(\widehat{\text{GATES}}(j) = |G_j|^{-1}\sum_{i \in G_j}(\hat{Y}_i(1) - \hat{Y}_i(0))\).
\textbf{Required:} \(\varepsilon\)-Fidelity for means; Joint for
ranking. \textbf{Validation:} L1, L4. \textbf{Classical:} Chernozhukov
et al.~(2018); requires ignorability + overlap + consistent CATE
estimator.

\hypertarget{d.3-family-iii-individual-level-and-distributional-effects}{%
\subsection{Family III: Individual-Level and Distributional
Effects}\label{d.3-family-iii-individual-level-and-distributional-effects}}

\hypertarget{d.3.1-individual-treatment-effect-ite}{%
\paragraph*{D.3.1 Individual Treatment Effect
(ITE)}\label{d.3.1-individual-treatment-effect-ite}}

\textbf{Definition.} \(\tau_i = Y_i(1) - Y_i(0)\). \textbf{DTCF
Estimator.} \(\hat{\tau}_i = \hat{Y}_i(1) - \hat{Y}_i(0)\).
\textbf{Required:} Joint Fidelity. \textbf{Validation:} L2 (supporting),
L4 (robustness). \textbf{Error decomposition:}
\(\text{MSE}(\hat{\tau}_i) = \mathbb{E}[(\hat{Y}_i(1) - Y_i(1))^2] + \mathbb{E}[(\hat{Y}_i(0) - Y_i(0))^2] - 2\text{Cov}(\cdot)\).
If errors independent across arms:
\(\text{MSE}(\hat{\tau}_i) \approx \text{MSE}_1 + \text{MSE}_0\).
\textbf{Classical:} Fundamentally unidentified.

\hypertarget{d.3.2-distribution-of-the-ite}{%
\paragraph*{D.3.2 Distribution of the
ITE}\label{d.3.2-distribution-of-the-ite}}

\textbf{Definition.} \(G_\tau(t) = \Pr(\tau_i \leq t)\). \textbf{DTCF
Estimator.}
\(\hat{G}_\tau(t) = N^{-1}\sum_i \mathbf{1}(\hat{\tau}_i \leq t)\).
\textbf{Required:} Joint Fidelity. \textbf{Validation:} L4.
\textbf{Bound:} \(\varepsilon_J + O(N^{-1/2})\). \textbf{Derived
quantities:} Variance, skewness, multimodality detection (Silverman's
test, dip test). \textbf{Classical:} Unidentified (Heckman, Smith \&
Clements, 1997). Partial identification by \citet{fan2010sharp}.

\hypertarget{d.3.3-probability-of-benefit-and-harm}{%
\paragraph*{D.3.3 Probability of Benefit and
Harm}\label{d.3.3-probability-of-benefit-and-harm}}

\textbf{Definition.} \(\pi_+ = \Pr(\tau_i > 0)\),
\(\pi_- = \Pr(\tau_i < 0)\). \textbf{DTCF Estimator.}
\(\hat{\pi}_+ = N^{-1}\sum_i \mathbf{1}(\hat{\tau}_i > 0)\).
\textbf{Required:} Joint Fidelity. \textbf{Validation:} L4.
\textbf{Bounds:} Via Theorem 8. \textbf{Classical:} Not identified;
bounded only.

\hypertarget{d.3.4-probability-of-causation}{%
\paragraph*{D.3.4 Probability of
Causation}\label{d.3.4-probability-of-causation}}

\textbf{Definition.} \(\text{PC} = \Pr(Y_i(0) = 0 \mid Y_i(1) = 1)\)
(binary outcomes). \textbf{DTCF Estimator.}
\(\widehat{\text{PC}} = \sum_i \mathbf{1}(\hat{Y}_i(1) = 1, \hat{Y}_i(0) = 0) / \sum_i \mathbf{1}(\hat{Y}_i(1) = 1)\).
\textbf{Required:} Joint Fidelity. \textbf{Validation:} L4.
\textbf{Classical:} Partially identified with wide bounds (Tian \&
Pearl, 2000). DTCF point estimate can be checked against these bounds as
consistency test.

\hypertarget{d.4-family-iv-mediation-effects}{%
\subsection{Family IV: Mediation
Effects}\label{d.4-family-iv-mediation-effects}}

\hypertarget{d.4.1-natural-direct-effect-nde}{%
\paragraph*{D.4.1 Natural Direct Effect
(NDE)}\label{d.4.1-natural-direct-effect-nde}}

\textbf{Definition.}
\(\text{NDE} = \mathbb{E}[Y_i(1, M_i(0)) - Y_i(0, M_i(0))]\).
\textbf{DTCF Estimator.}
\(\widehat{\text{NDE}} = N^{-1}\sum_i [\mathcal{S}_Y(X_i, 1, \hat{M}_i(0), U_i) - \mathcal{S}_Y(X_i, 0, \hat{M}_i(0), U_i)]\).
\textbf{Required:} Structural Fidelity. \textbf{Validation:} L1 per
component. \textbf{Classical:} Sequential ignorability \citep{imai2010identification}. \textbf{DTCF advantage:} Cross-world counterfactual directly
simulated.

\hypertarget{d.4.2-natural-indirect-effect-nie}{%
\paragraph*{D.4.2 Natural Indirect Effect
(NIE)}\label{d.4.2-natural-indirect-effect-nie}}

\textbf{Definition.}
\(\text{NIE} = \mathbb{E}[Y_i(1, M_i(1)) - Y_i(1, M_i(0))]\).
\textbf{DTCF Estimator.}
\(\widehat{\text{NIE}} = N^{-1}\sum_i [\mathcal{S}_Y(X_i, 1, \hat{M}_i(1), U_i) - \mathcal{S}_Y(X_i, 1, \hat{M}_i(0), U_i)]\).
\textbf{Consistency check:}
\(\widehat{\text{NDE}} + \widehat{\text{NIE}} = \widehat{\text{ATE}}\)
by construction.

\hypertarget{d.4.3-controlled-direct-effect-cde}{%
\paragraph*{D.4.3 Controlled Direct Effect
(CDE)}\label{d.4.3-controlled-direct-effect-cde}}

\textbf{Definition.}
\(\text{CDE}(m) = \mathbb{E}[Y_i(1,m) - Y_i(0,m)]\). \textbf{DTCF
Estimator.}
\(\widehat{\text{CDE}}(m) = N^{-1}\sum_i [\mathcal{S}_Y(X_i, 1, m, U_i) - \mathcal{S}_Y(X_i, 0, m, U_i)]\).
\textbf{Required:} Structural Fidelity (partial --- correct
\(F_{Y|D,M,X}\); does not require correct \(F_{M|D,X}\)).
\textbf{Classical:} Weaker than NDE --- requires
\(Y_i(d,m) \perp\!\!\!\perp D_i \mid X_i, M_i\).

\hypertarget{d.5-family-v-temporal-and-dynamic-effects}{%
\subsection{Family V: Temporal and Dynamic
Effects}\label{d.5-family-v-temporal-and-dynamic-effects}}

\hypertarget{d.5.1-time-varying-treatment-effects}{%
\paragraph*{D.5.1 Time-Varying Treatment
Effects}\label{d.5.1-time-varying-treatment-effects}}

\textbf{Definition.} \(\tau_i(t) = Y_i(1,t) - Y_i(0,t)\). \textbf{DTCF
Estimator.}
\(\hat{\tau}_i(t) = \mathcal{S}(X_i, 1, t, U_i) - \mathcal{S}(X_i, 0, t, U_i)\).
\textbf{Required:} Sequential Fidelity to horizon \(t\).
\textbf{Validation:} L1 per \(t\); L2 for trajectories.
\textbf{Classical:} Sequential ignorability; marginal structural models
(Robins et al., 2000).

\hypertarget{d.5.2-dynamic-treatment-regime-values}{%
\paragraph*{D.5.2 Dynamic Treatment Regime
Values}\label{d.5.2-dynamic-treatment-regime-values}}

\textbf{Definition.} \(V(\bar{g}) = \mathbb{E}[Y_i(\bar{g})]\); optimal
\(\bar{g}^* = \arg\max V(\bar{g})\). \textbf{DTCF Estimator.}
\(\hat{V}(\bar{g}) = N^{-1}\sum_i \hat{Y}_i(\bar{g})\); optimization via
grid search or backward induction. \textbf{Required:} Sequential
Fidelity. \textbf{Validation:} L1 per decision point; L3 if RCT data
available. \textbf{Classical:} g-computation, MSMs, Q-learning; all
require seq. ignorability + positivity. \textbf{DTCF advantage:} Any
regime evaluable by forward simulation, including never-implemented
ones.

\hypertarget{d.5.3-counterfactual-survival-functions}{%
\paragraph*{D.5.3 Counterfactual Survival
Functions}\label{d.5.3-counterfactual-survival-functions}}

\textbf{Definition.} \(S_d(t) = \Pr(T_i(d) > t)\);
\(\Delta_{\text{RMST}} = \int_0^{t^*}(S_1(t) - S_0(t))\,dt\).
\textbf{DTCF Estimator.}
\(\hat{S}_d(t) = N^{-1}\sum_i \mathbf{1}(\hat{T}_i(d) > t)\).
\textbf{Required:} \(\varepsilon\)-Fidelity on survival distributions.
\textbf{DTCF advantage:} No censoring --- simulator generates complete
trajectories.

\hypertarget{d.6-family-vi-spillover-and-interference-effects}{%
\subsection{Family VI: Spillover and Interference
Effects}\label{d.6-family-vi-spillover-and-interference-effects}}

\textbf{Definition.} Under partial interference (Hudgens \& Halloran,
2008): \(\text{DE}_i(d_{-i}) = Y_i(1, d_{-i}) - Y_i(0, d_{-i})\),
\(\text{SE}_i(d, d_{-i}, d_{-i}') = Y_i(d, d_{-i}) - Y_i(d, d_{-i}')\).
\textbf{DTCF Estimator.} If the simulator models interactions:
\(\widehat{\text{DE}}_i = \mathcal{S}(X_i, 1, d_{-i}, U_i) - \mathcal{S}(X_i, 0, d_{-i}, U_i)\).
\textbf{Required:} Multi-agent \(\varepsilon\)-Fidelity.
\textbf{Validation:} L1 on cluster outcomes; L3 with cluster-randomized
data. \textbf{Classical:} Cluster-level ignorability; exponential growth
in potential outcomes. \textbf{DTCF advantage:} Any allocation vector
evaluable computationally.

\hypertarget{appendix-e-joint-distribution-additional-mitigation-strategies}{%
\section{Joint Distribution --- Additional Mitigation
Strategies}\label{appendix-e-joint-distribution-additional-mitigation-strategies}}

\hypertarget{e.1-copula-sensitivity-analysis-full-protocol}{%
\subsection{Copula Sensitivity Analysis (Full
Protocol)}\label{e.1-copula-sensitivity-analysis-full-protocol}}

\textbf{Definition 4 (Copula Sensitivity Function).}
\(\psi_\theta(\rho) = \theta(C^{(\rho)}, F_1(\cdot \mid x), F_0(\cdot \mid x))\),
where \(C^{(\rho)}\) is a one-parameter copula family: - Gaussian:
\(C_{\text{Gauss}}^{(\rho)}(u,v) = \Phi_2(\Phi^{-1}(u), \Phi^{-1}(v); \rho)\).
- Frank:
\(C_{\text{Frank}}^{(\alpha)}(u,v) = -\alpha^{-1}\ln(1 + (e^{-\alpha u}-1)(e^{-\alpha v}-1)/(e^{-\alpha}-1))\).
- Clayton:
\(C_{\text{Clayton}}^{(\theta)}(u,v) = (u^{-\theta} + v^{-\theta} - 1)^{-1/\theta}\).

\textbf{Protocol 2.} (1) Validate marginals via L1. (2) Compute
\(\hat{\theta} = \psi_\theta(\hat{\rho}^*)\). (3) Compute
\(\psi_\theta(\rho)\) over grid \(\rho \in \{-0.9, -0.7, \ldots, 0.9\}\)
for at least one family. (4) Plot sensitivity function; report range.
(5) If flat: copula-robust. If sharp: flag as copula-dependent.

\hypertarget{e.2-structural-constraints-on-the-copula}{%
\subsection{Structural Constraints on the
Copula}\label{e.2-structural-constraints-on-the-copula}}

\textbf{Constraint 1 (Positive Dependence).} \(C_x(u,v) \geq uv\)
(positive quadrant dependence). Under PQD, \(\rho \geq 0\) and
\(\text{Var}(\tau_i) \in [(\sigma_1-\sigma_0)^2, \sigma_1^2 + \sigma_0^2]\).

\textbf{Constraint 2 (Stochastic Monotonicity).} \(Y_i(1) \geq Y_i(0)\)
a.s. implies \(\pi_- = 0\) and eliminates large portions of copula
space.

\textbf{Constraint 3 (Rank Invariance).} \(C_x = M\) (comonotonic);
joint fully determined by marginals; all Family III estimands identified
from marginals alone.

\textbf{Proposition 9.}
\(\Theta_{\text{constrained}} \subsetneq \Theta_{\text{FH}}\) whenever
\(\mathcal{C}_0 \subsetneq \mathcal{C}\).

\hypertarget{e.3-cross-validation-on-proxy-joints}{%
\subsection{Cross-Validation on Proxy
Joints}\label{e.3-cross-validation-on-proxy-joints}}

\textbf{Setting 1 (Repeated Measures).} Pre/post correlations provide
plausibility checks on the copula.

\textbf{Setting 2 (Crossover Trials).} Theorem 12: Fisher \(z\)-test
compares \(\hat{\rho}_{\text{sim}}\) to \(\hat{\rho}_{\text{obs}}\) from
crossover data.

\textbf{Setting 3 (Multi-Outcome Calibration).} Concordance matrix
discrepancy \(\Delta_R = \|\hat{R} - R\|_F\) across multiple observed
outcomes provides indirect copula evidence (Proposition 10).

\hypertarget{e.4-bayesian-copula-uncertainty}{%
\subsection{Bayesian Copula
Uncertainty}\label{e.4-bayesian-copula-uncertainty}}

Place prior \(\pi(C)\) over copulas (e.g., \(\rho \sim U(-1,1)\) for
Gaussian family). Update with proxy data via Theorem 12. For each
posterior draw \(\rho^{(s)}\), compute
\(\theta^{(s)} = \psi_\theta(\rho^{(s)})\). Report posterior mean and
credible interval.

This converts ``we don't know the copula'' from silence into quantified
uncertainty. For copula-robust estimands, the CI collapses regardless of
prior; for copula-sensitive estimands, width reflects genuine epistemic
uncertainty.

\hypertarget{appendix-f-llm-implementation-details}{%
\section{LLM Implementation
Details}\label{appendix-f-llm-implementation-details}}

\hypertarget{f.1-temperature-calibration}{%
\subsection{Temperature
Calibration}\label{f.1-temperature-calibration}}

Temperature \(T\) controls \(\hat{\sigma}_d^2(x)\). At \(T=0\):
deterministic (\(\hat{\sigma}^2 = 0\)). At high \(T\): over-dispersion.
Calibrate:
\[T^* = \arg\min_T \sum_{d,k} (\hat{\sigma}_d^2(\mathcal{X}_k; T) - \hat{\sigma}_{d,\text{obs}}^2(\mathcal{X}_k))^2.\]
One-dimensional optimization by grid search. The Funhouse Mirrors
mega-study \citep{peng2025funhouse} found LLM twins systematically under-dispersed,
suggesting \(T^* >\) default.

\hypertarget{f.2-scaling-and-computation}{%
\subsection{Scaling and
Computation}\label{f.2-scaling-and-computation}}

\(2N\) LLM calls (or \(N\) joint-prompt calls). Copula sensitivity
multiplies by grid size. For \(N > 10{,}000\): (i) batch inference; (ii)
covariate compression via clustering; (iii) surrogate modeling (LLM
generates training data, fast model handles inference); (iv) stratified
simulation (dense in high-CATE-variance strata).

\hypertarget{f.3-ethical-considerations}{%
\subsection{Ethical
Considerations}\label{f.3-ethical-considerations}}

\textbf{Consent.} Digital twin construction from personal data requires
consent or legal basis under GDPR/HIPAA. The twin's simulated status
does not exempt it.

\textbf{Right to non-simulation.} Individuals may object; ethical
frameworks must accommodate opt-out.

\textbf{Algorithmic fairness.} If the LLM encodes biases, DTCF estimates
are biased for those groups. Level 1 conditional calibration across
demographic strata detects differential fidelity.

\textbf{Transparency.} Decisions from DTCF estimates should disclose:
(i) counterfactuals are simulated; (ii) validation level achieved; (iii)
copula sensitivity; (iv) Fréchet-Hoeffding bounds.

\hypertarget{appendix-g-open-problems-and-future-directions}{%
\section{Open Problems and Future
Directions}\label{appendix-g-open-problems-and-future-directions}}

\textbf{Empirical validation.} Instantiate the DTCF in a specific domain
with real data and an RCT benchmark; demonstrate the full validation
protocol.

\textbf{Optimal simulator design.} Given fixed budget, what is the
optimal allocation between marginal accuracy and copula accuracy?

\textbf{Theoretical foundations for LLM fidelity.} Under what conditions
on training data and architecture does an LLM satisfy
\(\varepsilon\)-fidelity?

\textbf{Continuous treatments.} Extend to dose-response; validation
architecture must handle a continuum of doses; copula space becomes
infinite-dimensional.

\textbf{Multi-treatment comparisons.} \(K > 2\) treatments:
\(K\)-dimensional copula; Fréchet-Hoeffding bounds loosen; vine copulas
may be needed.

\textbf{Integration with classical methods.} Use the simulator as a
Bayesian prior, updated with observed data --- formally combining
structural knowledge with unbiasedness guarantees.

\end{document}